\documentclass[lettersize,journal]{IEEEtran}
\usepackage{amsmath,amsfonts}
\usepackage{algorithmic}
\usepackage{algorithm}
\usepackage{array}
\usepackage[caption=false,font=normalsize,labelfont=sf,textfont=sf]{subfig}
\usepackage{textcomp}
\usepackage{stfloats}
\usepackage{url}
\usepackage{verbatim}
\usepackage{graphicx}
\usepackage{cite}
\usepackage{booktabs}
\usepackage{multirow}
\usepackage{multicol}
\usepackage{colortbl}
\usepackage{amssymb}
\usepackage{mathtools}
\usepackage{adjustbox}
\usepackage{amsthm}
\usepackage{orcidlink}
\newtheorem{theorem}{Theorem}
\definecolor{mygrey}{rgb}{0.9,0.9,0.9}
\definecolor{gray}{rgb}{0.5,0.5,0.5}
\definecolor{darkred}{rgb}{0.9,0.1,0.1}
\newcommand{\reva}[1]{\textcolor{black}{#1}}
\newcommand{\rb}[1]{\textcolor{black}{#1}}

\definecolor{up_color}{RGB}{0, 128, 0}    % Dark Green for positive changes
\definecolor{down_color}{RGB}{255, 0, 0}  % 
\newcommand{\changeindicator}[1]{%
  \ifdim #1pt >0pt%
    \textcolor{up_color}{\tiny{\, +#1\%}}%
  \else%
    \textcolor{down_color}{\tiny{\, #1\%}}%
  \fi%
}

\begin{document}

\title{MoPE: Mixture of Prompt Experts for Parameter-Efficient and Scalable Multimodal Fusion}

\author{Ruixiang Jiang$^{\orcidlink{0000-0001-8666-6767}}$, Lingbo Liu$^{\orcidlink{0000-0001-8179-6685}}$, and Chang Wen Chen$^{\orcidlink{0000-0002-6720-234X}}$~\IEEEmembership{Fellow,~IEEE}

\thanks{Chang Wen Chen is the corresponding author.}
\thanks{Ruixiang Jiang and Chang Wen Chen are with the Department of Computing, The Hong Kong Polytechnic University, Hong Kong (email: rui-x.jiang@connect.polyu.hk, chen.changwen@polyu.edu.hk)}
\thanks{Lingbo liu is with the Research Institute of Multiple Agents and Embodied Intelligence, Pengcheng Laboratory, China (email: liulb@pcl.ac.cn)}
\thanks{Manuscript received May 15, 2025; revised Aug 23, 2025; Accepted Oct 09, 2025.}}

\markboth{MoPE: Mixture of Prompt Experts for Parameter-Efficient and
Scalable Multimodal Fusion}%
{Jiang \MakeLowercase{\textit{et al.}}: MoPE: Mixture of Prompt Experts for Parameter-Efficient and
Scalable Multimodal Fusion}

\IEEEpubid{0000--0000/00\$00.00~\copyright~2025 IEEE}

\maketitle

\begin{abstract}
Despite the demonstrated parameter efficiency of prompt-based fusion, its limited adaptivity and expressiveness hinder its effectiveness for multimodal applications at scale. In this paper, we present the first comprehensive study addressing these limitations. Our key motivation is to ``divide and conquer'' the vanilla prompt, traditionally shared across all instances, by generating instance-specific prompts. Specifically, we propose the Mixture of Prompt Experts (MoPE), a framework that significantly enhances prompt adaptivity and expressiveness by dynamically generating instance-specific prompts. MoPE leverages multimodal pairings as additional evidence, allowing the model to adaptively select optimal prompts tailored to each individual instance.  Unlike traditional prompt-fusion methods, which encounter scalability bottlenecks when optimizing long unified prompts, MoPE maintains fixed prompt length while effectively scaling the number of specialized experts.  Moreover, we investigate regularization terms to encourage expert specialization, resulting in highly adaptive and interpretable prompting. MoPE fundamentally changes the scaling dynamic, unlocking greater expressiveness and adaptability to complex multimodal relationships, enabling the model to selectively attend to task-relevant sub-sequences based on instance-specific multimodal input. Extensive experiments across six multimodal datasets spanning four modalities demonstrate state-of-the-art performance for multimodal fusion, matching or surpassing the performance of fine-tuning while requiring only 0.8\% of the trainable parameters. Code is available: \url{https://github.com/songrise/MoPE}.
\end{abstract}

\begin{IEEEkeywords}
Parameter-efficient multimodal fusion, Prompt tuning, mixture of experts, adaptivity and expressiveness.
\end{IEEEkeywords}

\section{Introduction}

\IEEEPARstart{R}{ecent} success of large-scale vision-language pretraining, such as CLIP~\cite{radford2021learning}, has demonstrated strong transferability to a wide range of vision-language tasks. However, extending the success of modalities-paired pretraining to downstream multimedia applications where modality may broadly span vision, language, audio, and beyond introduces unique challenges. Particularly when new data modalities or model architectures emerge, joint pretraining becomes computation-intensive and impractical due to the scarcity of paired data. In contrast, unimodal pre-trained models, such as ViT~\cite{dosovitskiy2020image}, Bert~\cite{devlin2019bert}, and Wav2Vec~\cite{baevski2020wav2vec}, are more readily available and offer greater flexibility, as they can be independently pre-trained, updated, and scaled. Therefore, to democratize pre-training to multimodal applications, a compelling question arises: 
 \begin{quote}
\textit{How can we efficiently combine separately pre-trained unimodal models for multimodal tasks?}
 \end{quote}

\begin{figure}
    \centering
    \includegraphics[width=\linewidth]{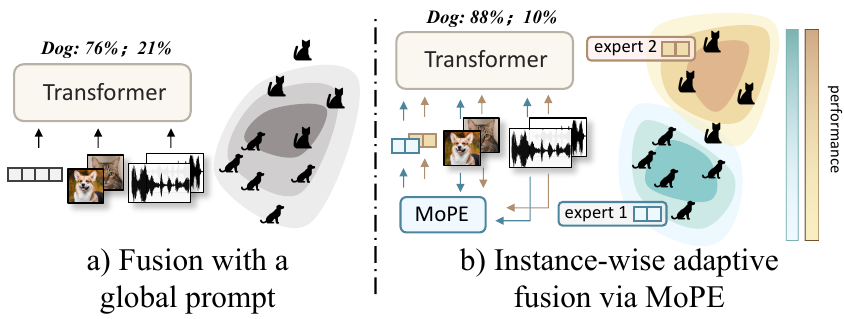}
\caption{\textbf{High-level motivation of MoPE-based multimodal fusion}. \rb{(a) Vanilla prompt fusion uses a non-adaptive global prompt (gray rectangles) for all inputs. It is challenging to fully capture the per-instance shift, resulting in suboptimal performance. (b) MoPE achieves instance-wise adaptivity by routing the most effective prompts for each instance. Its specialized expert effectively learns to divide the problem space into concept clusters. The prompt is then generated according to the predicted cluster, leading to a more expressive representation and superior performance.}}
    \label{fig:teaser}
\end{figure}

To address this challenge, recent progress in parameter-efficient multimodal fusion (PEMF), in particular, prompt fusion~\cite{li2021prefix,jia2022visual,zhou2022learning,liang2024querying,zhao2024copl} has offered promising solutions. Prompt fusion leverages unimodal pre-trained encoders while minimizing trainable parameters by freezing encoder weights and introducing a small set of learnable token embeddings called ``prompts''. These prompts fuse representations from different modalities and are fed into the transformer of a designated ``main'' modality encoder, enabling the reuse of pre-trained weights for multimodal processing. 

Despite its parameter efficiency, prompt fusion inherits a critical limitation from its underlying mechanism—prompt tuning: poor scalability. As widely observed~\cite{liang2022modular,tsimpoukelli2021multimodal,yang2022prompt,lester2021power}, prompts are effective in low-data regimes but fail to scale with increasing data or model size. Unlike fine-tuning or alternative parameter-efficient fine-tuning (PEFT) techniques such as LoRA~\cite{hu2021lora}, prompt-based methods plateau in performance, unable to effectively leverage additional resources. This scalability bottleneck significantly undermines prompt fusion's potential as an efficient solution for complex multimodal applications at scale.

\IEEEpubidadjcol

In this paper, we attribute the reduced scalability of prompt fusion to its limited \textbf{adaptivity} and \textbf{expressiveness}. Specifically, existing prompt fusion methods~\cite{tsimpoukelli2021multimodal,li2023efficient, liang2022modular} typically employ global prompts shared across all instances. As static, dataset-level priors, these prompts disregard the evidence present in instance-specific multimodal input, leading to a suboptimal representation for multimodal tasks~\cite{shi2023top,anderson2018bottom,lai2020understanding,sood2023multimodal}. Furthermore, the constrained expressiveness of prompt tuning (compared with fine-tuning) can lead to underfitting in multimodal datasets with long-tail distributions and complex cross-modal mappings~\cite{liu2021p,petrov2023prompting,wang2023universality}.

To overcome these limitations, a seemingly straightforward solution is to increase the number of learnable prompts, known as \textit{``length-scaling''}. Despite being a long-standing approach, the performance gains from length-scaling quickly reach saturation in both transfer-learning~\cite{lester2021power,xing2023dual} and fusion settings~\cite{liang2022modular,tsimpoukelli2021multimodal,yang2022prompt}. Furthermore, excessive prompt lengths can degrade performance~\cite{jia2022visual, yang2022prompt, khattak2023maple, li2021prefix,kim2023we,xing2023dual,hu2021lora}. Recent theoretical analyses have substantiated these empirical observations~\cite{wang2023universality, petrov2023prompting}, particularly regarding the difficulty in optimizing unified and long prompt vectors. Overall, evidence suggests that a structural redesign of prompt fusion is crucial for enhancing its scalability.

In this work, we present the \textbf{first} approach that simultaneously addresses the lack of adaptivity and expressiveness issues in prompt fusion. Our high-level idea is to divide-and-conquer the problem space utilizing multimodal pairing as an updated belief, as depicted in Figure~\ref{fig:teaser}. Specifically, we decompose the unified, long prompt into specialized and short prompts that are conditionally synthesized based on multimodal input. In the center of this decomposition is our proposed \textbf{M}ixture \textbf{o}f \textbf{P}rompt \textbf{E}xperts (MoPE) module. The MoPE module instance-wisely generates (\textit{i.e.,} routes) the most suitable prompts based on representation from all modalities. This architecture is inspired by the mixture-of-expert (MoE) designs~\cite{jacobs1991adaptive}, which gracefully mitigates the two challenges in scaling prompt fusion: (1) the multimodal routing mechanism ensures the effectiveness of generated prompts for each unique multimodal instance, thus significantly improving prompt adaptivity; and (2) increasing the number of experts (i.e., \textit{``expert-scaling''}) better scales the expressiveness compared with simply increasing the length of a unified prompt.  

To substantiate the effectiveness of MoPE, we conduct systematic experiments on a total of six datasets spanning four modalities, including image, text, audio, and video. Thanks to the improved adaptiveness and expressiveness, MoPE achieves superior performance and a higher parameter efficiency, compared to existing prompt fusion methods (e.g, \cite{li2023efficient,liang2022modular}). Compared to full-tuning, MoPE achieves on-par performance with only 0.8\% trainable parameters. To further understand the effect of introducing MoE into prompt fusion, we provide an extensive ablation analysis. Our analysis reveals that expert-scaling is more scalable than length-scaling, with monotonic performance gains and avoiding performance deterioration associated with overly-long prompts. MoPE also exhibits better scaling ability with respect to the size of the training data, whereas vanilla prompt fusion becomes less effective in full-shot training scenarios. Furthermore, by introducing regularization terms for expert routing, we observe the emergence of specialized prompt experts after end-to-end training. These specialized experts focus on particular concepts present in the dataset, which not only enhances prompt adaptivity but also provides a means for interpretable prompt fusion.

Our key contributions are summarized as follows:

\begin{itemize}
    \item To the best of our knowledge, we are the first to formulate the in-scalability issue of prompt fusion as a lack of adaptivity and expressiveness.
    \item We propose a novel instance-wise adaptive prompt decomposition technique to augment the adaptiveness of prompt fusion.
    \item We introduce the MoPE method for instance-wise dynamic prompt generation, which scales up the expressiveness of vanilla prompts.
    \item We study a combination of regularization terms to aid expert specialization in MoPE. 
    \item Extensive experiments on six datasets spanning four modalities demonstrate state-of-the-art performance and parameter efficiency for prompt-based multimodal fusion.
\end{itemize}

\section{Related Works}

\subsection{Prompt Tuning for Transfer Learning} 

Prompt tuning~\cite{li2021prefix, lester2021power} is a PEFT method that learns continuous token embeddings to condition frozen pre-trained models.  It was initially popularized in the natural language processing (NLP) community for transfer learning with Transformers, aiming to automate the manual prompt engineering. Subsequently, it is gradually introduced to solve computer vision~\cite{jia2022visual, bahng2022exploring,liu2022prompt,jiang2023clip,zhu2023prompt,zhou2024hcvp} and audio~\cite{liu2024audio} tasks. Recent studies have also applied it to adapt multimodal models~\cite{zhou2022learning,khattak2023maple,zeng2023temporally,chen2023attentive}.

The effectiveness of prompt tuning is widely demonstrated across various modalities and tasks~\cite{jia2022visual, bahng2022exploring,liu2022prompt,khattak2023maple,chen2025improving}. \rb{Particularly for low-shot model adaptation~\cite{xing2023dual,jiang2023clip,zhou2022learning}, it often achieves comparable or superior performance to full fine-tuning, with only approximately 1\% trainable parameters. However, this strength diminishes as the training data or trainable parameter scales~\cite{li2021prefix,yang2022prompt,hu2021lora}, leading to a noticeable performance gap with full fine-tuning and other PEFT methods (e.g., LoRA~\cite{hu2021lora}).} At the core of this challenge is the in-scalability of prompt-tuning: the conventional length-scaling paradigm quickly reaches performance saturation, and even leads to deterioration~\cite{jia2022visual, yang2022prompt, khattak2023maple, li2021prefix,kim2023we,xing2023dual}. \rb{Recent theoretical analyses~\cite{petrov2023prompting,wang2023universality} further underpinned the limited expressiveness of length-scaling. Inspired by these findings, we seek a more effective scaling paradigm with prompts by introducing a MoE-like architecture.}

\subsection{Prompt Fusion} Apart from model adaptation, prompting can also fuse separately pre-trained \textit{unimodal models} for multimodal tasks. Frozen~\cite{tsimpoukelli2021multimodal} first introduced a method where the visual representation is mapped as a few input tokens to query frozen language models (LMs). PromptFuse and BlindPrompt~\cite{liang2022modular} improved upon this by introducing tunable prompts to the LM for cross-modal alignment. PICa~\cite{yang2022empirical} translates images into discrete text captions to prompt frozen LMs. These methods treat tokens from different modalities independently and lack explicit cross-modal interaction. Recently, PMF~\cite{li2023efficient} introduced an interactive prompt fusion method for the vision and text modalities, building on the strong assumption that both encoders are white-box and have the same Transformer architecture. CoPL~\cite{zhao2024copl} targets audio-visual tasks and proposes collaborative prompts for efficient fusion. \rb{Nearly all of these methods are extensions of the standard prompt tuning paradigm, and as such, they inherit its inherent limitations in scalability. Our work presents the first systematic analysis aimed at resolving this issue, identifying a lack of adaptivity and expressiveness in the fusion mechanism as the key bottleneck. In addressing these challenges, our proposed design not only enhances performance but also offers superior modularity, making it compatible with heterogeneous and black-box backbone architectures.}

\subsection{MoE with Transformers} The MoE is widely recognized as a computationally efficient scaling paradigm for large models, including Transformers~\cite{riquelme2021scaling,mustafa2022multimodal}. The MoE layers scale up the model parameter by increasing the number of experts, which are typically implemented as parallel feed-forward networks. A router is trained within each MoE layer to direct each token embedding to the most suitable expert(s). \rb{Existing literature predominantly employs a sparse router to reduce computational overhead, which selects only a small subset of experts for each token~\cite{lepikhin2020gshard,qin2021learning,riquelme2021scaling,shazeer2017outrageously}. However, the discrete, non-differentiable nature of the top-$k$ selection operation used in sparse routing can introduce training instabilities. This has motivated recent explorations into soft routing as a differentiable alternative~\cite{puigcerver2023sparse}, which has found applications in diverse contexts such as neuro-imaging~\cite{quan2024psychometry}.} Overall, MoE represents a scaling paradigm orthogonal to plain parameter scaling. Moreover, the routing mechanism effectively equips models with enhanced instance-adaptivity by partially activating model weights~\cite{jacobs1991adaptive,zhang2021enhancing}. \rb{Concurrent with our research, several studies have explored combining MoE with prompt tuning for tasks such as semantic understanding~\cite{wu2024mixture}, continual learning~\cite{le2024mixture}, and automatic prompt engineering~\cite{wang2024one}. In contrast to these efforts, our work is distinctly focused on applying the MoE framework to address the fundamental limitations in adaptivity and expressiveness that are inherent to prompt-based multimodal fusion.}

\section{Method}
Our high-level objective is to efficiently fuse unimodal pre-trained models for downstream multimodal tasks via prompting. To achieve this goal, we tackle the key technical challenges of prompt fusion: limited expressiveness and adaptivity.

In this section, we begin by reviewing the mechanism and the limitations of vanilla prompting in Section~\ref{sec: prelim}. In Section~\ref{sec: cpt}, we introduce our instance-wise adaptive prompt decomposition, which serves as the basis of our fusion pipeline. Section~\ref{sec: mop} introduces the proposed MoPE in detail. Finally, we introduce the regularization methods to aid MoPE training in Section~\ref{sec: reg}.

\begin{figure*}[t]
    \centering
    \includegraphics[width=.99\linewidth]{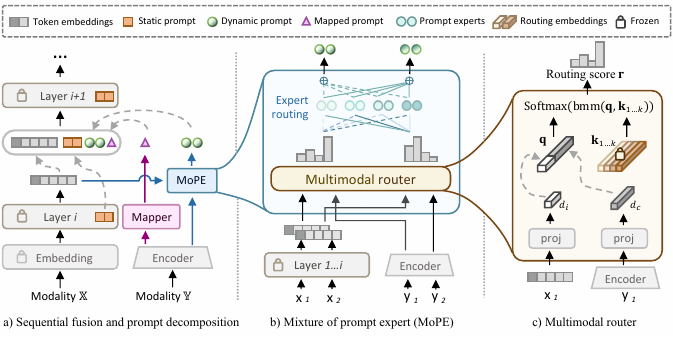}
    \caption{\textbf{Architecture overview.} \textbf{(a)} A sequential fusion pipeline is employed, where the representation from complementary modality $\mathbb{Y}$ guides the prompting of modality $\mathbb{X}$. Three types of prompts are used at each layer, which are concatenated to the token embeddings. \textbf{(b)} MoPE is introduced to generate the dynamic prompt, which routes the most effective dynamic prompt based on multimodal representations. Here, $(x_1,y_1), (x_2,y_2)$ indicate two input pairs; \textbf{(c)} Inside the multimodal router, we project the representation from each modality to get cross-modal and inter-modal embeddings. The concatenated embeddings $\mathbf{q}$ are used to query the routing embedding $\mathbf{k}_{1\dots k}$ paired with each expert for routing score calculation. Better viewed with color. }
    \label{fig:arch}
\end{figure*}

\subsection{Preliminary: Limitation of Vanilla Prompt Tuning}\label{sec: prelim}

Consider the case of using a Transformer~\cite{vaswani2017attention} or its variants~\cite{dosovitskiy2020image} to extract features from an embedded input sequence $\mathbf{x}^0 \in \mathbb{R}_{s\times d_x}$, where $s$ is the sequence length and $d_x$ is the embedding dimension. Prompt-tuning freezes all pre-trained parameters and optimize a small number of continuous embeddings (\textit{i.e.,} prompts) $\mathbf{P}\in \mathbb{R}_{l\times d}$ concatenated to the input of each layer\footnote{In this paper, we use the term ``prompt tuning'' to refer to the prompts applied to all Transformer layers.}, where $l$ is the length of the prompt vector and usually $d=d_x$. The input of the $i$-th layer layer $L^i$ could be denoted as:

\begin{equation}
    \mathbf{\hat{x}}^i = [x^{i-1}_0,\mathbf{P},\mathbf{T}^{i-1}],
    \label{eq:vanprompt}
\end{equation}
where $x^{i-1}_0\in \mathbb{R}_{d_x}$ denotes the \verb|[CLS]| token, $\mathbf{T}^{i-1}\in \mathbb{R}_{s\times d_x}$ is the token embedding from the previous layer, and $[, ]$ denotes the concatenation operation. \rb{The variable sequence length $s$ is typically handled by the encoder itself.} \reva{We call $\mathbf{P}$ as a globally-shared prompt, as it indiscriminately applies to \rb{all input instances}. This strategy suffers from two key limitations:}  

\textbf{Limited Adaptivity.} Top-down attention mechanism~\cite{shi2023top,anderson2018bottom,lai2020understanding,sood2023multimodal} is crucial for multimodal tasks such as visual question answering~\cite{lin2024evaluating} (VQA) and visual entailment~\cite{xie2018visual} (VE). These tasks require models to selectively attend to different sub-sequences according to multimodal input.  \rb{However, vanilla prompt fusion struggles to fully capture this adaptivity, as the global prompt indiscriminately applies to every instance. Similar concern on this technical issue appears in recent literature~\cite{yang2024dgl}, bringing a call for adaptive prompt generation~\cite{bai2024soft,chi2025learning,loedeman2022prompt,wang2022learning,wang2022dualprompt}. Importantly, this lack of adaptivity can provably result in sub-optimal task performance compared to fully adaptive approaches. Specifically, as analyzed in Appendix~\ref{sec: theory}, when framing the objective of prompting into function approximation of per-instance optimal attention patterns, global prompts are almost always less effective than the instance-wise adaptive ones. We address this limitation by introducing instance-wise adaptive prompts via MoPE.}

\textbf{Limited Expressiveness.} It has been recently proved that prompting have strictly limited expressiveness~\cite{petrov2023prompting,wang2023universality}, characterized as function approximation error.  \rb{Theoretically, it is mainly due to the working mechanism of prompt, which can only \textbf{bias} the pre-trained attention pattern~\cite{petrov2023prompting}, whereas LoRA~\cite{hu2021lora} and finetuning have the potential to learn new attention patterns. This unique characteristic of prompting is persistent even with $l \to \infty$. While it avoids catastrophic forgetting, it also imposes a strict upper bound on its expressiveness. In practice, however, the empirical gains from length-scaling are reportedly lower than the gains predicted by theoretical analysis. This gap is primarily attributed to a ``competing optimization scheme'' that arises when training a long prompt vector ($l>1$)~\cite{petrov2023prompting}. In other words, although a longer prompt provides more trainable parameters, its optimization is significantly more challenging, and a poorly optimized prompt can degrade performance rather than improve it. This issue is a key insight into the ineffectiveness of length-scaling reported throughout the prompt-tuning literature~\cite{jia2022visual, yang2022prompt, khattak2023maple, li2021prefix,kim2023we,xing2023dual,hu2021lora}. In this paper, we are motivated to address the performance degradation associated with length-scaling in order to narrow the gap between its empirical and theoretical expressiveness.}

\subsection{Instance-wise Adaptive Prompt Decomposition} \label{sec: cpt}
\rb{To fully exploit the multimodal interplay for adaptive prompting, we instance-wisely condition the prompting of one modality on the other(s). To this end, a sequential fusion pipeline is employed.} Specifically, let $x\in\mathbb{X}, y\in\mathbb{Y}$ be a pair of multimodal inputs, and $\mathcal{E}_{\mathbb{X}}$,  $\mathcal{E}_\mathbb{Y}$ be the encoder of each modality. \rb{One may choose a specific modality to guide the other depending on task intrinsic, such as using the text modality to guide the image encoding for visual question answering. This leads to a directed fusion, where the encoding of \textit{main modality} $\mathbb{X}$ is additionally conditioned on the representation of \textit{complementary modality(ies)} $\mathbb{Y}$}. 

Building upon the sequential pipeline, we decompose the vanilla prompt vector $\mathbf{P}$ used in $\mathcal{E}_{\mathbb{X}}$ into three types of \textit{specialized prompts} $[\mathbf{P}_s, \mathbf{P}_d, P_m]$. The static prompt $\mathbf{P}_s \in \mathbb{R}_{l\times d_x}$ is a globally-shared prompt vector indiscriminately applied to all instances. The dynamic prompt $\mathbf{P}_d\in \mathbb{R}_{l\times d_x}$, however, is adaptive to different instances. To synthesize it, we first encode global-level features from the complementary modality $\psi_y = \mathcal{E}_\mathbb{Y}(y) \in\mathbb{R}_{d_y}$. Then, a MoPE module $R(\cdot,\cdot)$ is used, which takes as input the representation from all modalities and outputs the dynamic prompt. For the mapped prompt $P_m\in\mathbb{R}_{d_x}$, we apply a lightweight mapper $f_m(\cdot)$ to map the complementary feature into a single prompt. This prompt injects fine-grained cross-modal information that complements the dynamic prompt. In summary, the input of layer $L^i$ of $\mathcal{E}_{\mathbb{X}}$ becomes:

\begin{equation}
    \mathbf{\hat{x}}^i = [x^{i-1}_0, \underbrace{\mathbf{P}_s,R(x_0^{i-1},\psi_y), f_m(\psi_y)}_{\text{Decomposed prompts}},\mathbf{T}^{i-1}]
\end{equation}

The whole process is illustrated in Figure~\ref{fig:arch}-(a). \rb{Effectively, the decomposed prompts capture specialized roles for prompt fusion: global model adaptation, per-instance model adaptation, and fine-grained cross-modal information fusion. This design features diverse representation and is broadly related to the multiple knowledge representation~\cite{yang2021multiple}. }

\subsection{Mixture of Prompt Experts}\label{sec: mop}

\rb{To scale up the expressiveness of prompts, we decompose the unified long prompt $\mathbf{P}_d$ via a MoPE module in each layer to effectively scale up prompt fusion, as illustrated in Figure~\ref{fig:arch}-(b).} Concretely, a MoPE module consists of a multimodal router, $k$ prompt experts and their associated \textit{routing embeddings} $\{(\mathbf{E}_j,\mathbf{k}_j)\}_{j=1}^k$  where $\mathbf{E}_j\in\mathbb{R}_{l\times d}$ is an expert, $\mathbf{k}_j\in\mathbb{R}_{d_r}$ is the routing embedding associated with the expert, and $d_r\ll d_x $ is the dimension of routing embedding.

For every Transformer layer, we route each instance based on representations of all modalities, as visualized in Figure~\ref{fig:arch}-(c). For example, in the two modality setting, the router is parameterized as two layer-specific linear transformations $\mathbf{W}_y^{i}\in\mathbb{R}_{d_y\times d_c}$ and $\mathbf{W}_x^{i}\in\mathbb{R}_{d_x\times d_i}$, where $d_c$, $d_i$ are the dimension of cross-modal and inter-modal routing embedding, respectively, and $d_i + d_c = d_r$. The cross-model embedding is projected from $\psi_y$, while inter-model embedding is mapped from the global-level feature in the preceding layer of main modality encoder $L^{i-1}$ (i.e., $x_0^{i-1}$). Finally, we concatenate both embeddings to get a multimodal query embedding $\mathbf{q}\in\mathbb{R}_{d_r}$, used to query the routing score $\mathbf{r}\coloneqq[r_1,\cdots, r_j]$ associated with each expert:
\begin{equation}
\label{eq:route_score}
    \mathbf{q} =[\psi_y\mathbf{W}_y^{i}, x_0^{i-1}\mathbf{W}_x^{i} ], \quad
    r_j = \frac{\operatorname{exp}(\mathbf{q}^\top\mathbf{k}_j/\tau + \epsilon)}{\sum_{n=1}^k \operatorname{exp}(\mathbf{q}^\top\mathbf{k}_n/\tau + \epsilon)},
\end{equation}
where $\tau=0.1$ is the temperature hyper-parameter, and $\epsilon\sim\mathcal{N}(0,\frac{1}{k^2})$ is sampled noise~\cite{riquelme2021scaling}. When there is more than one complementary modality, the additional embedding could be easily extended by learning additional projections. The dynamic prompt is obtained \rb{by a soft routing mechanism, which is a convex combination of all experts at this layer}:

\begin{equation}
\mathbf{P}_d = \sum_{j = 1}^k r_j \mathbf{E}_j
\label{eq:route}
\end{equation}

% \rb{The routing scheme (Discussion here)}

\begin{figure*}
    \centering
    \includegraphics[width=\linewidth]{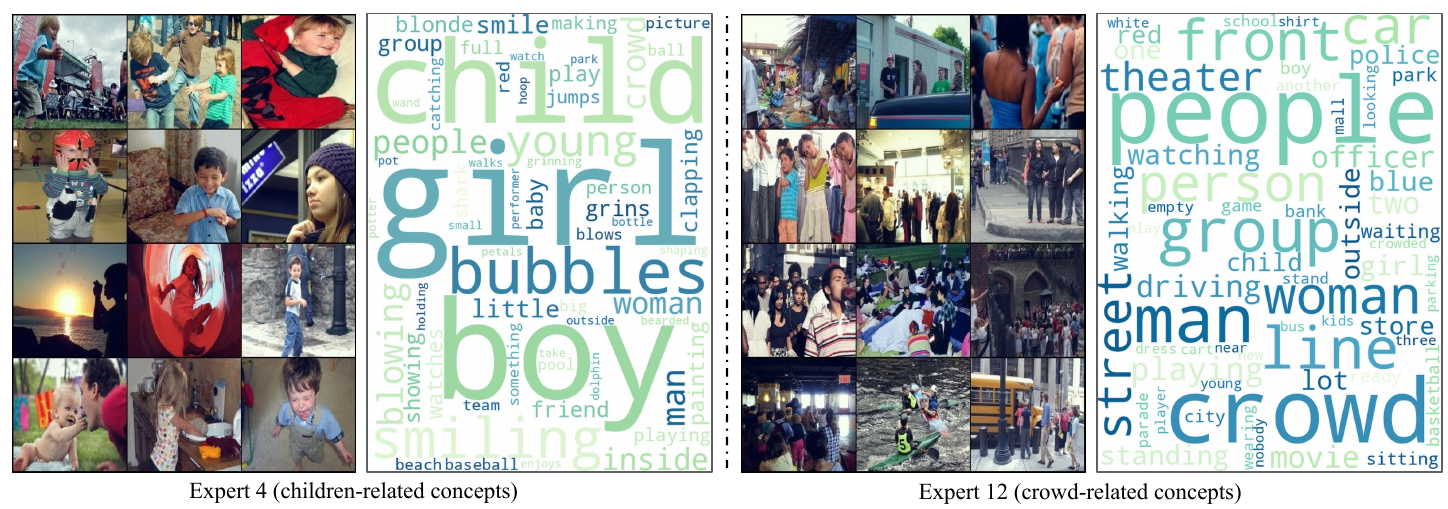}
    \caption{\textbf{\rb{Illustration of expert specialization in MoPE.}}  \rb{ Each expert learns to specialized in a group of concepts, akin to a soft cluster in the semantic space. For instance, Expert 3 activates for prompts related to children and playful activities, while Expert 12 activates for scenes involving crowds and public spaces.}}
    \label{fig:expert}
\end{figure*}

\subsection{Regularizing Expert Routing} \label{sec: reg}

\rb{During the training of MoPE, we empirically observe that a few experts dominate the routing decisions across the majority of input instances.} This phenomenon is also reported in prior MoE literature~\cite{eigen2013learning,shazeer2017outrageously}. Such domination prevents the utilization of most experts, which significantly limits the adaptivity and expressiveness of the trained model. \rb{This section introduces two strategies designed to prevent this degenerate routing behavior and promote effective expert specialization.}

\textbf{Frozen Routing Embeddings.} \rb{As defined in Equation~\ref{eq:route_score}, the routing mechanism involves two sets of parameters: the multimodal router $(\mathbf{W}_x, \mathbf{W}_y)$ and the expert routing embeddings ${\mathbf{k}_{1...j}}$. To achieve adaptive routing, optimization should focus on the router, as it is conditioned on the input, rather than on the static expert embeddings. We observed that when both components are learnable, the expert embeddings tend to be updated more rapidly than the router. This leads to a degenerate routing scenario where the experts that happen to be initialized with higher scores receive disproportionately larger gradient signals throughout training, creating a "rich-get-richer" dynamic. To address this, we freeze the expert embeddings after an orthogonal initialization~\cite{saxe2013exact}. This strategy prevents any expert from having an unfair initial advantage and ensures that routing decisions are learned adaptively based on the input.}
 
% \textbf{Frozen Routing Embedding.} \rb{Routing as in Equation~\ref{eq:route_score} involves two sets of parameters: multimodal router $(\mathbf{W}_x, \mathbf{W}_y)$ and the routing embeddings ${\mathbf{k}_{1...j}}$. To achieve adaptive routing, the router instead of the routing embeddings shall be optimized, since these embeddings are not input-conditioned. When both are learnable, we observe the routing embeddings will be updated more rapidly before the router. This results in degenerated routing, where the expert that happen to be initiated with higher scores receive disproportionately larger gradient signals during throughout training. To tackle this issue, we freeze the routing embeddings with orthogonal initialization~\cite{saxe2013exact} to avoid unbalanced advantage of experts during training.}
 
 \textbf{Importance Loss.} To further aid expert specialization, we add an additional importance loss~\cite{shazeer2017outrageously,riquelme2021scaling} to penalize dominant experts. For a batch of input $\mathbf{B} = \{(x_1,y_1),(x_2,y_2),\cdots, (x_b,y_b)\}$, the importance of expert-$j$ is defined as the summed routing score in this batch
 
 \begin{equation}
     \operatorname{Imp}(\mathbf{E}_j)= \sum_{(x,y) \in \mathbf{B}} r_j
 \end{equation}
 The importance loss is calculated as the mean coefficient of variation of all experts' importance averaged across all layers:

\begin{equation}
\begin{aligned}
    \mathcal{L}_{imp} &= \sum_{All\ layers} \operatorname{\sigma}\left( \left(\frac{\operatorname{std}(\{\operatorname{Imp}(\mathbf{E}_j)\}_j^k)}{\operatorname{mean}(\{\operatorname{Imp}(\mathbf{E}_j)\}_j^k)} \right)^2; \gamma \right),
\end{aligned}
\end{equation}
\reva{where $\operatorname{\sigma}(\cdot)$ truncates the loss to zero when the coefficient of variation is less than a threshold parameter $\gamma = 0.1$.} 

This importance loss, originally designed to balance computational budgets in conventional MoE models~\cite{shazeer2017outrageously,riquelme2021scaling}, is repurposed here to explicitly discourage the over-domination of certain experts. The additional threshold constraint $\gamma$ is necessary due to instance-wise routing of MoPE, which typically has a smaller effective sample size than per-token routing, and hence naturally exhibits higher variance.

\section{Experiments}
\subsection{Dataset and Tasks}
\textbf{UPMC Food-101}~\cite{wang2015recipe} is a comprehensive multimodal dataset designed for fine-grained recipe classification.  The dataset contains 90,840 image-text pairs for 101 food classes. 

\textbf{MM-IMDB}~\cite{arevalo2017gated} is a multimodal movie classification dataset.  It comprises 25,956 pairs of images and texts, each pair including a movie poster and a plot summary. The task is multi-label classification across a spectrum of 23 movie genres with a long-tail distribution.

\textbf{SNLI-VE}~\cite{xie2019visual} is a large-scale multimodal dataset with 565,286 image-text pairs. The task for this dataset is visual entailment, which requires the model to decide whether a hypothesis matches the given premise. This dataset provides image and text premises, while the hypothesis is always in text modality. \reva{Following prompt fusion literature~\cite{li2023efficient}, we only consider the image premise, meaning that the input to the model is image premise + text hypothesis.}

\textbf{MUStARD}~\cite{castro2019towards} is a dataset for multimodal sarcasm detection. \reva{The raw dataset contains 690 video clips with an even number of sarcastic and non-sarcastic labels. Each video is provided with annotations in the text modality, describing the speaker and dialog in that video.} Collectively, three modalities are available: video, audio extracted from the video, and text. \reva{We experiment with the speaker-dependent setting, meaning that the identity of the author is available during inference.}

\textbf{RefCOCO} and \textbf{RefCOCO+}~\cite{kazemzadeh2014referitgame} are two datasets for referring expression comprehension, \reva{and in this paper we consider the referring expression segmentation (RES) task, where the model shall predict a 2D binary mask according to natural language expression.} RefCOCO contains referring expressions of any type, while 
RefCOCO+ features expressions that do not contain object positions.

Details on the data processing are in Appendix~\ref{sec: data_detail}.

\subsection{Experiment Setups}
\textbf{Metrics.} The metric on SNLI-VE, UPMC Food-101 is accuracy (\%), MM-IMDB is F1-Macro and F1-Micro, and on MuSTARD is precision (\%). Mean Intersection over Union (mIoU) is the metric for RefCOCO and RefCOCO+.

\textbf{Architecture Details.} In main experiment, we use publicly-available pre-trained Swin-B~\cite{liu2021swin} as the image encoder \reva{with an input resolution of $224\times224$, except for RES task, where the resolution is $384\times384$}. We use pre-trained Bert-base-uncased~\cite{devlin2019bert} as the text encoder, and pre-trained Wav2Vec2~\cite{baevski2020wav2vec} as the audio encoder. For video encoding, we follow PromptFuse~\cite{liang2022modular} to use a pre-trained ViT~\cite{dosovitskiy2020image} for extracting image features from $n=8$ evenly sampled frames, the temporally-averaged global feature is used to represent the whole video. 

Image is treated as the main modality unless otherwise specified. Following the experiment setup in~\cite{jia2022visual,li2023efficient, liang2022modular}, we finetune dataset-specific heads. Specifically, linear classification heads are used for all classification tasks, and we use a standard UperNet~\cite{xiao2018unified} head for segmentation tasks. We implement the mapper as a two-layer MLP with GeLU nonlinearity. Regarding the prompt, we set $l=6$ and use $k=4$ experts by default, which strikes a balance between performance and parameter size. The prompts are applied to all layers of the main modality encoder. Vanilla prompt tuning~\cite{li2021prefix} with $l^\prime = 4$ is used to tune the $\mathcal{E}_\mathbb{Y}$. Further implementation details can be found in the Appendix~\ref{sec: more_detail}.

\begin{table*}[t]
\centering
\caption{\textbf{Quantitative results on multimodal classification.} Our method achieves the best performance and parameter-efficiency against all prompt-fusion methods. I: Image, L: Language, A: Audio, V: Video. The main modality is \underline{underlined}, and the best prompt tuning method is in \textbf{bold}, and (\dag): Our re-implementation with prompt applied to all layers.}
\resizebox{\linewidth}{!}{%
\begin{tabular}{llccccccc}
\toprule
& & & &  \multicolumn{3}{c}{\underline{I}+L} & A+\underline{L} & A+V+\underline{L} \\
\cmidrule(lr){5-7} \cmidrule(lr){8-8} \cmidrule(lr){9-9}
&Method & Param  & Speed (ms) & SNLI-VE &UPMC Food& MM-IMDB & MUsTARD & MUsTARD \\
& & & & Acc \% ($\uparrow$) & Acc \% ($\uparrow$) & F1-macro/F1-micro ($\uparrow$) & Pre \% ($\uparrow$) & Pre \% ($\uparrow$) \\
\midrule
\multirow{5}{*}{\rotatebox[origin=c]{90}{\textit{fine-tuning}}} & ImgOnly & 86.9M & 16.45$\pm$0.2 & 33.34 & 75.64 & 39.21/53.85 & - & - \\
&TextOnly & 109.0M & 7.28$\pm$0.9 & 69.58 & 86.92 & 58.80/65.37 & 65.41 & 65.41 \\
&LateConcat & 196.0M & 22.79$\pm$0.9 & 72.01 & 93.19 & 60.43/67.77 & 68.82 & 69.40 \\
&SeqFuse & 197.0M  & 23.32$\pm$1.2 & 74.28 & 93.73 & 59.22/66.34 & 68.13 & 71.35 \\
&MMBT & 196.5M & 15.91$\pm$1.2 & 67.58 & 94.10 & 60.80/66.10 & - & - \\
\midrule
\multirow{11}{*}{\rotatebox[origin=c]{90}{\textit{prompt-tuning}}} & P-ImgOnly & 0.1M  & 21.01$\pm$1.1 & 33.34 & 76.65 & 33.70/50.04 & - & - \\
&P-TextOnly & 0.1M  & 7.57$\pm$1.2 & 64.86 & 81.01 & 52.19/61.16 & 59.27 & 59.27 \\
&P-LateConcat & 1.3M  & 28.59$\pm$2.1 & 64.29 & 90.27 & 56.95/64.23 & 60.71 & 65.17 \\
&P-SeqFuse & 1.1M  & 28.52$\pm$1.9 & 67.01 & 81.27 & 55.57/63.98 & 63.72 & 65.73 \\
&P-MMBT & 0.9M  & 16.71$\pm$1.3 & 67.58 & 81.07 & 52.95/59.30 & - & - \\
&PromptFuse & \textless0.1M  & 28.75$\pm$0.9 & 64.53 & 82.21 & 48.59/54.49 & 63.76 & 64.20 \\
&BlindPrompt & \textless0.1M  &  29.52$\pm$1.2 & 65.54 & 84.56 & 50.18/56.46 & 62.01 & 63.80 \\
&PromptFuse(\dag) & 0.1M  & 29.26$\pm$1.7 & 64.94 & 82.14 & 50.78/60.96 & 64.73 & 66.29 \\
&PMF & 2.5M  & 32.21$\pm$0.8 & 71.92 & 91.51 & 58.77/64.51 & - & - \\
&\cellcolor{mygrey}Ours (ViT, $k\!=\!4$) & \cellcolor{mygrey}1.6M & \cellcolor{mygrey}23.17$\pm$1.1 & \cellcolor{mygrey}73.47 & \cellcolor{mygrey}91.55 & \cellcolor{mygrey}\textbf{62.37}/\textbf{68.73} & \cellcolor{mygrey}- & \cellcolor{mygrey}- \\
&\cellcolor{mygrey}Ours (Swin, $k\!=\!4$) & \cellcolor{mygrey}1.6M & \cellcolor{mygrey}30.47$\pm$1.4 & \cellcolor{mygrey}73.14 & \cellcolor{mygrey}91.54 & \cellcolor{mygrey}61.93/68.19 & \cellcolor{mygrey}67.12 & \cellcolor{mygrey}67.35 \\
&\cellcolor{mygrey}Ours (Swin, $k\!=\!16$) & \cellcolor{mygrey}2.6M & \cellcolor{mygrey}30.44$\pm$1.3 & \cellcolor{mygrey}\textbf{73.59} & \cellcolor{mygrey}\textbf{92.05} & \cellcolor{mygrey}\textbf{62.01/68.24} & \cellcolor{mygrey}\textbf{68.73} & \cellcolor{mygrey}\textbf{69.94} \\
\bottomrule
\end{tabular}
}

\label{tab: classfication}
\end{table*}

\begin{table}[htbp]
\centering
\caption{\textbf{Quantitative result on referring image segmentation.} We report the total number of parameters in million (including both the encoder and the segmentation head), and metrics (mIOU) on RefCOCO and RefCOCO+.}
\label{tab:segmentation_results}
\resizebox{\columnwidth}{!}{
\begin{tabular}{c@{\hskip 0.5cm}c@{\hskip 0.5cm}ccc|ccc}
\toprule
\multirow{2}{*}{Method} & \multirow{2}{*}{Param} & \multicolumn{3}{c}{RefCOCO, mIOU ($\uparrow$)} & \multicolumn{3}{c}{RefCOCO+, mIOU ($\uparrow$)} \\
\cmidrule{3-5} \cmidrule{6-8}
       &       & val & testA&testB & val & testA&testB \\
\midrule
SeqFuse & 231.0M & 53.48 & 55.76 & 52.03& 40.22 & 42.2& 37.91 \\
P-SeqFuse & 35.1M & 47.69 & 46.23 & 45.81 & 30.66 & 31.48& 28.79 \\
PromptFuse(\dag) & 35.1M & 43.23 & 39.71 & 47.74 & 27.72 & 33.37& 23.67 \\

\midrule
Ours & 35.5M & \textbf{58.40} & \textbf{60.03} & \textbf{53.23} & \textbf{43.80} & \textbf{46.12} & \textbf{38.88}\\
\bottomrule
\end{tabular}}

\end{table}

\textbf{Training Details.} 
All models are trained for a maximum of 10 epochs with early stopping, using the AdamW~\cite{loshchilov2017decoupled} optimizer with a learning rate of $4\times10^{-4}$ for the main modality and $5 \times 10^{-4}$ for the complementary modality. All models are trained with a single RTX-4090 GPU. \reva{The convergence takes approximately 0.6 hr, 5 hrs, 2.5 hrs, 0.3 hr, and 7 hrs on MM-IMDB, SNLI-VE, UPMC Food-101, MUStARD, and RefCOCO(+), respectively.}

\subsection{Compared Methods.}    
We mainly compare our methods with existing open-sourced prompt-based fusion methods, including MMBT~\cite{kiela2019supervised}, Frozen~\cite{tsimpoukelli2021multimodal}, PromptFuse and BlindPrompt~\cite{liang2022modular}, and PMF~\cite{li2023efficient}. Among these methods, the setting in PromptFuse~\cite{liang2022modular} is the most similar to ours, as it also assumes one modality to be a black-box and is by design compatible with more than two modalities. While CoPL~\cite{zhao2024copl} and QaP~\cite{liang2024querying} are two related prompt-fusion methods, we exclude them from our primary comparison due to differences in evaluation datasets and the unavailability of their code at the time of writing. To ensure a fair comparison, all prompt-tuning baselines utilize the same prompt length ($l=6$) as our method.

In addition to those methods, the following baselines are considered: 

    \textit{ImgOnly / TextOnly}. Fine-tune one encoder only, and the input of the other modality is discarded.

    \textit{P-ImgOnly / P-TextOnly}. Only prompt-tune one encoder.

    \textit{LateConcat}.  This baseline fine-tunes both encoders, concatenates their features, and learn a linear head for classification.

    \textit{P-LateConcat}. Similar to \textit{LateConcat} but prompt-tune each encoder instead of fine-tuning.

    \textit{SeqFuse}. This baseline first extracts features from the complementary modality and maps them to the embedding space of the main modality encoder by an MLP. Both encoders are fine-tuned. This is a strong baseline that can be considered as our method without MoPE, but with all parameters fine-tuned.

    \textit{P-SeqFuse}. Similar to \textit{SeqFuse} but prompt-tune each encoders. 

\subsection{Quantitative Comparisons}

\textbf{Multimodal Classification.} The quantitative results of multimodal classification with all methods are summarized in Table~\ref{tab: classfication}. We also list the GPU wall-clock time (with batch size $b=16$, in milliseconds), and number of trainable parameters (in millions) of the full model for each method.

Our method outperforms all prompt-based fusion methods and is competitive with fine-tuning. Specifically, when compared with the fine-tuning baselines, \textit{SeqFuse} and \textit{LateConcat}, our method delivers competitive accuracy on the UPMC Food-101 dataset and superior results on the SNLI-VE, MM-IMDB, and MUStARD datasets (V+L), while requiring as few as 0.8\% of the trainable parameters. This demonstrates the strong parameter-efficiency of MoPE, which efficiently leverages the rich knowledge in pre-trained unimodal models for multimodal tasks.

The proposed method also outperforms \textbf{all} existing prompt fusion methods, including PromptFuse~\cite{liang2022modular}, BlindPrompt~\cite{liang2022modular}, and PMF~\cite{li2023efficient}. Notably, our method ($k=4$) outperforms the current SOTA, PMF, by a significant margin on all datasets while requiring $37\%$ fewer parameters. In particular, the PMF lacks an effective scaling mechanism. MoPE addresses the scaling limitation by utilizing the parameters in a more effective way, bringing stronger performance. Moreover, our method \reva{is more flexible and modular than PMF, as PMF is designed for modalities with homogeneous Transformer architecture (e.g., Bert and ViT). It is non-trivial to extend PMF for tasks that require heterogeneous architectures (e.g., CNN and Transformer) or with more than two modalities, such as the MUStARD dataset.} \rb{By contrast, our method is compatible with heterogeneous architectures and tasks with more than two modalities, such as experiments on MUsTARD.}

\textbf{Referring Expression Segmentation.} \rb{Prior prompt fusion methods have focused on classification tasks. This focus is not coincidental, which we argue is due to a representational bottleneck. Their low-capacity fusion may be sufficient for global classification but fails to produce the rich, spatially-aware features that dense prediction tasks like RES demand. As a result, simply attaching a new prediction head is ineffective if the underlying features from the backbone lack the necessary expressiveness.}

\rb{Our MoPE is designed to overcome this limitation by enhancing the expressiveness and adaptivity of prompt-tuned backbone. As a proof-of-concept in Tab.~\ref{tab:segmentation_results}, MoPE significantly outperforms all compared methods on RES. Crucially, it surpasses the fully fine-tuned \textit{SeqFuse} baseline using only 15.3\% of its parameters, demonstrating superior expressiveness without the catastrophic forgetting that plagues full fine-tuning. This experiment confirms MoPE's effectiveness for challenging dense prediction tasks.}

\begin{figure*}[!ht]
    \centering
    \includegraphics[width=\linewidth]{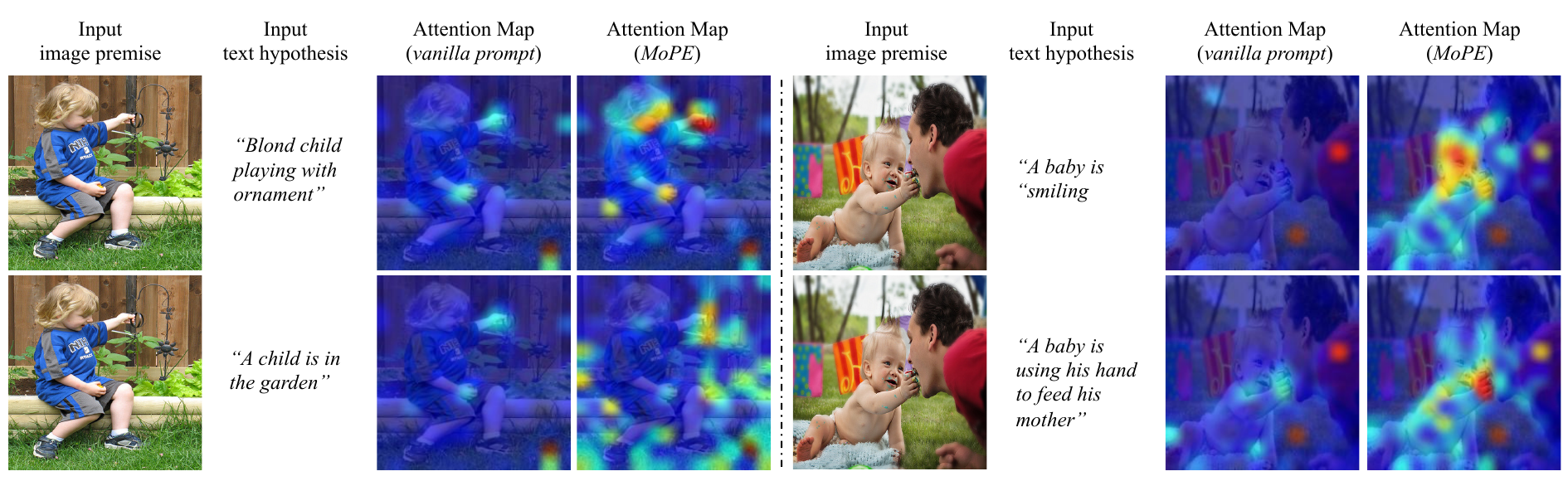}
    \caption{\textbf{Visualization of attention maps produced by different prompting methods.} \rb{Compared with the method with vanilla prompt that produces a similar attention map for each instance, the attention map produced by MoPE is adaptive. It grounds more attentively to the most-related tokens according to the text query, fulfilling the requirement of top-down attention.} Better viewed with color, warm colors indicate a higher attention score.}
    \label{fig:attn}
\end{figure*}

\subsection{\rb{Visualization of Expert Specialization}}\label{sec:route_vis} \rb{Through end-to-end training, prompt experts in MoPE spontaneously specialize in focusing on different ``concept clusters''. These clusters are not a single concept or a few words, but semantically meaningful and related features in a high-dimensional representation space. Furthermore, by weighting these experts via routing (i.e., Equation~\ref{eq:route}), MoPE achieves an even higher expressiveness by mixing different concept clusters to fully explore the semantic space.}

\rb{Figure~\ref{fig:expert} provides a qualitative illustration of this behavior. Specifically, we recorded the routing score of each instance on a fine-tuned MoPE model. We then visualized the raw inputs of instances with a high score for each expert. By doing so, we can get an intuitive understanding of the concept cluster that an expert specializes in. For example, we observe that Expert-4 has learned a subspace related to children, while Expert-12 focuses on crowds. The divergent focus of these two experts demonstrates clear evidence of specialization. Furthermore, it also provides an interpretable way of understanding the soft prompting, which is a clear advantage over black-box prompt generation methods such as \cite{zhou2022learning,chi2024adapting,bai2024soft,loedeman2022prompt}. To better understand the behavior of this expert specialization, we provide additional visualization in Appendix~\ref{sec:vqa}.}

\subsection{Visualization of Attention Maps}
MoPE demonstrates greater adaptivity compared to vanilla prompts for multimodal fusion. To illustrate this, we visualize the attention maps of the last layer of the vision encoder fused by MoPE and vanilla prompt (i.e., \textit{P-SeqFuse}). As shown in Figure~\ref{fig:attn}, the attention map with MoPE more effectively localizes to the corresponding region specified by the text condition. In contrast, the attention generated by the vanilla prompt appears indiscriminative across different text inputs, exhibiting a typical non-adaptive bottom-up attention pattern rather than a top-down one. This indicates that vanilla prompts lack instance-wise adaptivity, which is a crucial requirement for multimodal understanding.

\section{Analysis and Discussion}

\begin{table}[!t]
    \caption{\textbf{Ablation on prompt decomposition.} Our full method with all types of prompts achieves the best result.}
    \centering
\resizebox{\columnwidth}{!}{
    \begin{tabular}{c|ccc}
    \toprule
         Prompt & SNLI-VE Acc($\uparrow$)  & UPMC Food Acc($\uparrow$) & MM-IMDB F1($\uparrow$) \\ \midrule
        $[\mathbf{P}_s]$ & $33.34\textcolor{gray}{\scriptsize{\pm .01}}$ & $76.65\textcolor{gray}{\scriptsize{\pm .07}}$ & $33.70\textcolor{gray}{\scriptsize{\pm .55}}/50.04\textcolor{gray}{\scriptsize{\pm .27}}$ \\
$[\mathbf{P}_d]$ & $64.26\textcolor{gray}{\scriptsize{\pm .41}}$ & $74.79\textcolor{gray}{\scriptsize{\pm .38}}$ & $46.54\textcolor{gray}{\scriptsize{\pm .77}}/59.71\textcolor{gray}{\scriptsize{\pm .35}}$ \\
$[P_m]$ & $33.47\textcolor{gray}{\scriptsize{\pm .32}}$ & $73.06\textcolor{gray}{\scriptsize{\pm .12}}$ & $24.84\textcolor{gray}{\scriptsize{\pm .14}}/45.10\textcolor{gray}{\scriptsize{\pm .32}}$ \\
$[\mathbf{P}_s, \mathbf{P}_d]$ & $66.76\textcolor{gray}{\scriptsize{\pm .26}}$ & $75.13\textcolor{gray}{\scriptsize{\pm .14}}$ & $49.09\textcolor{gray}{\scriptsize{\pm .43}}/60.89\textcolor{gray}{\scriptsize{\pm .37}}$ \\
$[\mathbf{P}_s, P_m]$ & $65.01\textcolor{gray}{\scriptsize{\pm .18}}$ & $81.27\textcolor{gray}{\scriptsize{\pm .22}}$ & $55.57\textcolor{gray}{\scriptsize{\pm .63}}/63.98\textcolor{gray}{\scriptsize{\pm .35}}$ \\
$[\mathbf{P}_d, P_m]$ & $71.39\textcolor{gray}{\scriptsize{\pm .59}}$ & $91.21\textcolor{gray}{\scriptsize{\pm .16}}$ & $60.15\textcolor{gray}{\scriptsize{\pm .37}}/67.14\textcolor{gray}{\scriptsize{\pm .17}}$ \\\midrule
$[\mathbf{P}_s, \mathbf{P}_d, P_m]$ & $\mathbf{73.59}\textcolor{gray}{\scriptsize{\pm .15}}$ & $\mathbf{92.05}\textcolor{gray}{\scriptsize{\pm .11}}$ & $\mathbf{62.01}\textcolor{gray}{\scriptsize{\pm .21}}/\mathbf{68.24}\textcolor{gray}{\scriptsize{\pm .12}}$ \\
        \bottomrule
    \end{tabular}
}

    \label{tab:ablate_prompt}
\end{table}

\subsection{Our Decomposed Prompts are Collaborative}

We first ablate all types of prompts in our instance-wise adaptive prompt decomposition, with results across three random seeds reported in Table~\ref{tab:ablate_prompt}. Our full method achieves the best performance, which indicates that the prompts are collaborative.

Specifically, (1) adding the dynamic prompt yields significant gains of 13.5\%, 13.26\%, 8.4\%/6.5\% on the three datasets. \reva{This is achieved by instance-wise conditioning the prompt with cross-modal interaction, which is fundamentally different from previous prompt-fusion methods~\cite{liang2022modular,tsimpoukelli2021multimodal} where there is no explicit interaction of prompts. (2) While the dynamic prompt is also conditioned on $\psi_y$, it cannot replace the mapped prompt, as they have different responsibilities. This is due to the dynamic prompt having a low-rank bottleneck dimension of $ d_i + d_c$, which significantly undermines the fine-grained information from the complementary modalities. By contrast, the mapped prompt retains the high-rank cross model information with a mapper $f_m(\cdot)$. (3) The static prompt is also necessary, as it could be interpreted as a special expert always being routed to capture global-level features, akin to a bias term.}

\begin{figure}[t]
    \centering
    \includegraphics[width=\linewidth]{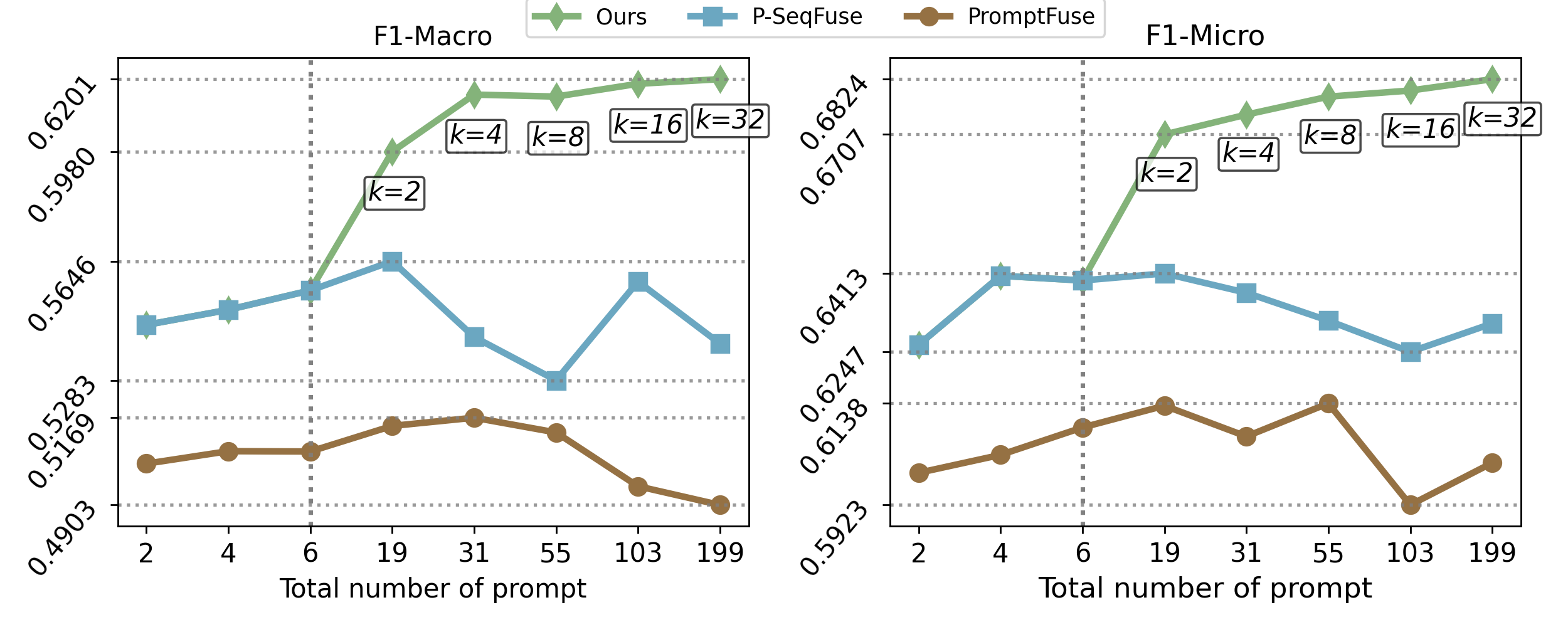}
    \caption{\textbf{More experts v.s. longer prompts.} We compare increasing the number of experts, $k$, versus lengthen prompt, $l$. Expert-scaling consistently outperforms length-scaling, exhibiting a linear growth trend.  Conversely, length-scaling suffers from deterioration with long prompts.}
    \label{fig:ablate_k}
\end{figure}

\subsection{Expert-scaling is More Expressive than Length-Scaling} This subsection compares the proposed expert-scaling with length-scaling. Our starting point for MoPE is $l=6$ prompts and $k = 2$ experts (when $k \le 1$, the MoPE degenerates into a vanilla prompt), which account for a total of $(2+1)\times6+1 = 19$ tunable prompt embeddings per layer. We increase the number of $k$, and at each step, we also report the result of increasing $l$ in the vanilla prompt to the same total number of prompts. The results are summarized in Figure~\ref{fig:ablate_k}.

Our findings indicate expert-scaling is more effective than length-scaling. Specifically, adding the MoPE design with as few as $k=2$ experts leads to a noticeable performance enhancement. More importantly, as we progressively increase the number of experts, the performance continues to improve in a monotonic manner. This consistent enhancement contrasts sharply with the behavior observed in length-scaling, where increasing the prompt length does not yield proportional performance gains and is often accompanied by performance degradation, which is also observed in prior studies~\cite{jia2022visual, yang2022prompt, khattak2023maple, li2021prefix, xing2023dual}. \rb{This experiment validates that expert scaling narrows the gap between empirical and theoretical expressiveness, fulfilling the objective described in Section~\ref{sec: prelim}. }

\begin{figure}
    \centering
    \includegraphics[width=\linewidth]{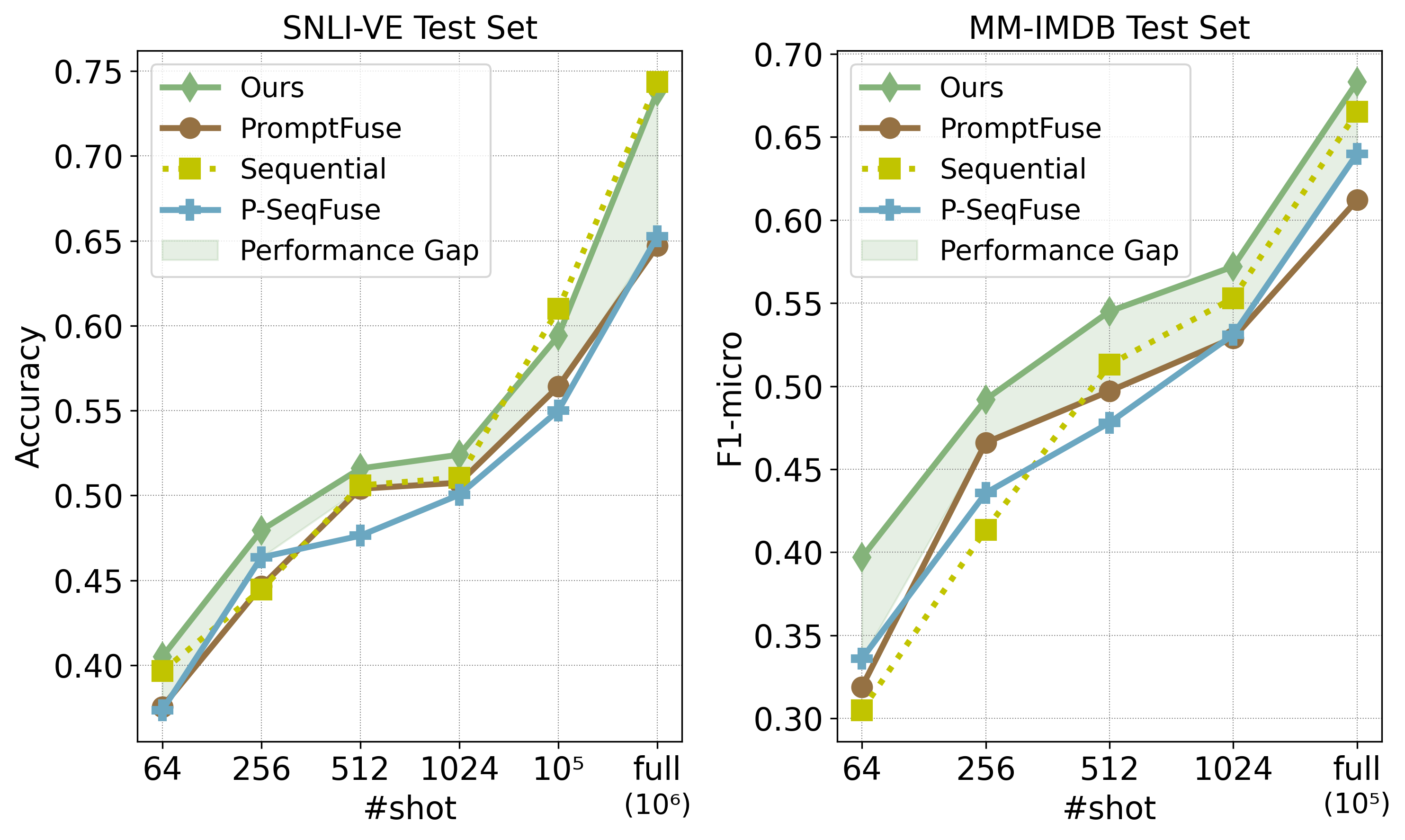}
    \caption{\textbf{Scaling performance with increased training data}. We show the performance of our method and representative methods as we progressively increase the amount of training data, or ``\#shots''. Our method outperforms other prompt fusion methods at all data scales. The performance gap between the best compared prompt fusion method is shaded.}
        \label{fig:n_shot}
\end{figure}

\subsection{MoPE Demonstrates Superior Data Scalability} It has been observed that the performance of existing prompt-based methods does not scale effectively with the volume of training data~\cite{liang2022modular,tsimpoukelli2021multimodal,yang2022prompt}. To evaluate MoPE's scalability, we designed an experiment measuring performance across a range of data regimes. We then train MoPE and other prompt-tuning and fine-tuning methods on the same subset of SNLI-VE and MM-IMDB.

Our MoPE-based method demonstrates superior scalability compared to other prompt fusion methods, as demonstrated in Figure~\ref{fig:n_shot}. Specifically, we consistently match the results of the fine-tuning method, \textit{SeqFuse}, on the SNLI-VE dataset, while surpassing it on the MM-IMDB dataset. By contrast, the methods based on vanilla prompts, PromptFuse, and \textit{P-SeqFuse} are less scalable with respect to increased training data, resulting in a consistent performance gap between our method and the compared prompt fusion methods. This performance disparity becomes more pronounced on larger datasets, such as when training with $10^5$ shots or using the full training set. This result directly demonstrates the effectiveness of MoPE in scaling up the expressiveness of prompt fusion, overcoming the limitations of vanilla prompts.

\begin{figure}[t]

    % \begin{minipage}[t]{0.55\linewidth} % Right minipage for the figure
        % \centering
        \includegraphics[width=\linewidth]{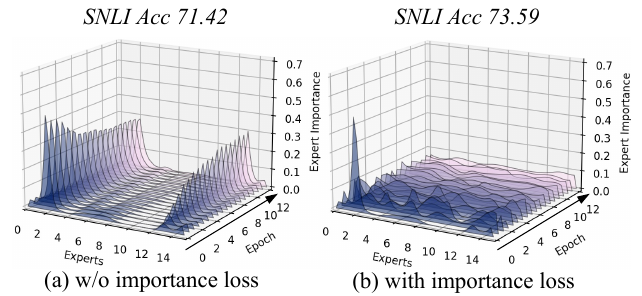}
        \caption{\textbf{\rb{Effect of the importance loss}}. \rb{ We visualize how the importance (z-axis) of all experts (x-axis) in the last Transformer layer changes during training (y-axis). (a) Without importance loss, only a few experts are used throughout training. (b) The importance loss ensures balanced utilization of all experts, leading to stronger performance.}}
        \label{fig:route-time}
    % \end{minipage}
\end{figure}

\subsection{\rb{Modularity of MoPE}} \rb{A key advantage of the MoPE framework is modularity and architectural agnosticism. Specifically, MoPE does not impose constraints on the architecture of cross-modal encoders $\mathcal{E}_\mathbb{Y}$, as the MoPE routing network only requires a fixed-dimension feature vector from the complementary modality. This makes it fully compatible with diverse architectures like Transformers or Convolutional Neural Networks (CNNs).} 

\rb{In this section, we empirically validate this flexibility on the MUStARD dataset by replacing the Wav2Vec2 audio encoder with a pretrained, frozen CNN-based model PANNs~\cite{kong2020panns}, with only an additional trainable linear projection to align feature dimensions. Table~\ref{tab:cmp_cnn} presents the result of this ablation. Notably, the result with PANNs \textbf{outperforms} the Wav2Vec-based encoder. The experiment demonstrated that MoPE not only integrates seamlessly with the CNN but also achieves a substantial performance improvement. This result confirms that MoPE is a truly model-agnostic framework, allowing it to be paired with the optimal specialist encoder to maximize downstream performance.}

\begin{table}[!t]
    \caption{\textbf{\rb{MoPE are Compatible with Different Architecture}}. \rb{We compare the performance of the Transformer (Wav2Vec2) and CNN (PANNs) backbones for audio encoding on the MusTARD dataset.}}
    \centering
    \begin{tabular}{ccc}
    \toprule
         Backbone  & MUsTARD (A+\underline{L}) $\uparrow$ & MUsTARD (A+V+\underline{L}) $\uparrow$  \\\midrule
         Wav2Vec2~\cite{baevski2020wav2vec} & 67.12 & 67.35\\
         PANNs~\cite{kong2020panns} &\textbf{72.47}&\textbf{75.28}\\
         \bottomrule
    \end{tabular}

    \label{tab:cmp_cnn}
\end{table}

\subsection{\rb{The Importance Loss Aids Adaptive Routing}}
The importance loss is crucial for avoiding degenerate routing solutions. In Figure~\ref{fig:route-time}, we visualize how the importance (i.e., average routing score) of each expert changes during training. Without the importance loss, routing adheres to its initial state, resulting in a skewed distribution where a few experts are always being routed throughout the optimization process. Introducing the importance loss encourages expert specialization by penalizing highly unbalanced expert importance distributions. \rb{As a result, the model achieves stronger adaptivity by utilizing the entire pool of experts to generate the dynamic prompt, leading to a substantial performance improvement.}

\begin{table*}[!htbp]
    \centering
    \caption{\rb{\textbf{Performance Comparison of Hard (Top-$n$) and Soft Routing.} \rb{The proposed soft routing outperforms the alternative hard routing, while maintaining the same computation efficiency.}}}
    \resizebox{0.9\linewidth}{!}{%
    \begin{tabular}{llcccccc}
    \toprule

     &Routing&  Param  & Speed & SNLI-VE &UPMC Food& MM-IMDB &RefCOCO (val)  \\
    & & & ms & Acc \% ($\uparrow$) & Acc \% ($\uparrow$) & F1-macro/F1-micro ($\uparrow$) & mIoU (\%)  \\

    \midrule 

    $k$=4 & Hard (Top-1)&  1.6M &  30.33$\pm$1.1 &  72.03  &  89.53 &  61.25/67.86 & 51.71   \\
    $k$=4 & Soft&  1.6M &  30.47$\pm$1.4 &  73.14 &  91.54 &  61.93/68.19 &  54.33   \\
    \midrule
    $k$=16 & Hard (Top-1)&  1.6M &  30.37$\pm$1.0 &  73.14 &  91.54 &  61.44/68.42 & 51.53   \\
    $k$=16 & Hard (Top-3)&  1.6M &  30.65$\pm$0.9 &  72.75  &  90.94 &  61.42/68.30 & 52.84    \\
    $k$=16 & Soft&  1.6M &  30.44$\pm$1.3 &  \textbf{73.59} &  \textbf{92.05} &  \textbf{62.01/68.24} &\textbf{58.40}  \\
    \bottomrule
    \end{tabular}
    }
    \label{tab: cmp_hardroute}
    \end{table*}

\subsection{\rb{Extended Analysis on MoPEs Routing}}

\rb{\textbf{Routing Sparsity.} First, we study the effect of different routing mechanisms (i.e., Equation~\ref{eq:route_score}). We compare the proposed soft (dense) routing with the hard (sparse) routing mechanism. To achieve dense routing, we keep all the designs of MoPE but only apply an additional Top-$n$ gate on the calculated routing score $\operatorname{Top}(\mathbf{r},n)$. We experiment with different total numbers of experts $k$ and the number of selected experts for the hard routing. Table~\ref{tab: cmp_hardroute} summarizes the results for this comparison.}

\rb{The results show that the soft routing mechanism is consistently stronger than the alternative hard routing, without additional computation overhead. The performance gap is especially noticeable on relatively large datasets (e.g., SNLI-VE) and challenging tasks (e.g., RES). This suggests that soft routing is empirically more expressive than the hard one with a prompt-based expert, despite the latter being favored in previous MoE literature for efficiency considerations~\cite {lepikhin2020gshard,fedus2022switch}. Our findings of applying soft routing are also broadly in line with recent MoE literature~\cite{quan2024psychometry,puigcerver2023sparse}.}

\rb{\textbf{Dimension of Routing Embeddings.} In addition to analyzing routing sparsity, we study the dimensions of the routing embeddings. Specifically, we investigate how the cross-modal dimension ($d_c$) and the intra-modal dimension ($d_i$) should be configured. Table~\ref{tab:ablate_d} presents the results of this ablation study. The results indicate that a high cross-modal dimension ($d_c > d_i$) combined with a non-zero intra-modal dimension ($d_i > 0$) yields the strongest performance. This suggests that for routing the dynamic prompt, the cross-modal representation ($\psi_y$) is more informative than the intermediate intra-modal feature ($x_0^{i-1}$).}

\rb{Our finding that $d_c > d_i$ and $d_i > 0$ is optimal aligns with the design principle outlined in Section~\ref{sec: cpt}, which posits that the task intrinsics should determine the fusion strategy. This finding is intuitive, as in many typical multimodal tasks, one modality (e.g., text) usually guides the encoding of another. For instance, in RES, the vision model should attend to different image regions based on the text query, a mechanism known as top-down attention~\cite{lin2024evaluating,anderson2018bottom}. A larger cross-modal dimension ($d_c$) better facilitates this instance-wise adaptivity. Conversely, a lower $d_c$ would cause the vision encoding to behave more like bottom-up attention. Furthermore, a non-zero intra-modal dimension ($d_i > 0$) is also beneficial, as it prevents routing decisions from becoming entirely dependent on the complementary modality.
}

\begin{table}[!htbp]
    \centering
    \caption{\textbf{\rb{Ablation on Cross-model Embedding dimension $d_c$ and Inter-model Embedding Dimension $d_i$.}} \rb{$d_c > d_i$ and $d_i>0$ results in best performance in general.}}
    \resizebox{\columnwidth}{!}{
    \begin{tabular}{c|ccccc}
        \toprule
          Dataset &$d_i=10$, $d_c=0$& $d_i=8$, $d_c=2$ & $d_i=5$, $d_c=5$ & $d_i=2$, $d_c=8$ &$d_i=0$, $d_c=10$  \\
        \midrule

        MM-IMDB($\uparrow$) & 54.50/64.63& 60.12/67.02& 61.22/67.51 & \textbf{62.01/68.24} & 61.02/67.93 \\
        UPMC Food($\uparrow$) &89.55& 90.52& 91.77&92.05&\textbf{92.11}\\
        SNLI-VE($\uparrow$) & 68.30&69.89&72.01&\textbf{73.59}& 71.98 \\
        \bottomrule
    \end{tabular}
    }

    \label{tab:ablate_d}
\end{table}

\begin{table}[!h]
    \caption{\rb{\textbf{Result of frozen routing embedding}. Frozen routing embedding is slightly better than the learned one, and orthogonal initialization slightly outperforms others.}}
    \centering
    \resizebox{\linewidth}{!}{
            \begin{tabular}{c|cccc}
            \toprule
            Routing Embed & SNLI-VE Acc ($\uparrow$) & UPMC  Food Acc ($\uparrow$) & MM-IMDB F1 ($\uparrow$) \\\midrule
                    Learned & 72.13 & 91.25 & 60.71/67.69 \\
            Frozen (Normal) & 73.21 & 91.54 & 61.00/67.47 \\            Frozen (Xavier) & 72.25 & 91.64 & 61.22/67.23 \\
           Frozen (Orthogonal) & \textbf{73.59} & \textbf{92.01} & \textbf{62.01}/\textbf{68.25} \\     
            \bottomrule
            \end{tabular}}

            \label{tab:frozen}
\end{table}

\rb{\textbf{Ablation on Frozen and Orthogonal Initialization.}  Finally, we ablate the design choice of frozen and orthogonal routing embeddings. Table~\ref{tab:frozen} presents the results of this ablation. The proposed frozen and orthogonal initialization offers the strongest performance. This performance gain is achieved by avoiding competing optimization with the router, resulting in more adaptive routing.}

\subsection{\rb{Extended Analysis on Time and Space Complexity}}
\label{sec:complexity}

\rb{In this section, we study the time and space complexity of expert-scaling in MoPE. Specifically, we measure (1) the cumulative inference time of the routing mechanism across all layers, and (2) the GPU memory consumption required for storing all experts during training. To achieve this, we monitor the model using the PyTorch profiler with batch size (batch $b=16$) and varying number of experts ($k$). We report both the raw time and memory values, as well as their proportion relative to a complete forward pass of the entire model. The results of this analysis are presented in Table~\ref{tab:complexity_analysis}.}

\begin{table}[!htbp]
\centering
\caption{\rb{\textbf{Time and space complexity of MoPE as the number of experts ($k$) increases}. Overhead is measured in absolute terms (microseconds/\textmu s and Megabytes/MiB) and as a percentage of a full forward pass of the model.}}
\label{tab:complexity_analysis}
\begin{tabular}{lcccccc}
\toprule
& \multicolumn{5}{c}{\textbf{Number of Experts ($k$)}} \\
\cmidrule(lr){2-7}
\textbf{Metric} & \textbf{4} & \textbf{8} & \textbf{16} & \textbf{32} & \textbf{64} & \textbf{128} \\
\midrule
Time (\textmu s) & 251 & 254 & 263 & 410 & 556 & 834 \\
Time (\%) & 0.8\% & 0.8\% & 0.8\% & 1.3\% & 1.8\%  & 2.6\% \\
\midrule
Mem (MiB) & 3.26 & 3.26 & 3.27 & 3.31 & 3.37 & 3.97 \\
Mem (\%) & 0.2\% & 0.2\% & 0.2\% & 0.2\% & 0.2\% & 0.2\%\\
\bottomrule
\end{tabular}
\end{table}

\rb{As shown in the table, the computational overhead introduced by MoPE is minimal across both time and space dimensions. The routing operation's latency scales efficiently, and even with 128 experts, it accounts for only 2.6\% of the total forward pass time. This high efficiency is due to our routing mechanism (Equation~\ref{eq:route_score}), which scales linearly ($\mathcal{O}(k)$) with the number of experts. This is much more efficient than length-scaling, whose time complexity is $\mathcal{O}(l^2)$, as all prompts are involved in the self-attention.  Similarly, the memory footprint for storing the expert prompts is also negligible. Storing 128 distinct experts requires only 3.97 MiB, consuming a mere 0.2\% of the total GPU memory used by the model.}

\rb{This analysis empirically confirms that MoPE is a highly efficient framework. It enables a substantial increase in model capacity and flexibility with only a marginal increase in computational cost, following the principles of parameter-efficient fusion.}

\subsection{\rb{Comparison with Other PEFT Methods}}

\begin{table}[!t]
    \centering
    \caption{\rb{\textbf{Comparison With Other PEFT Methods.} MoPE achieves the best results with a smaller number of trainable parameters.}}
    \begin{adjustbox}{width=\columnwidth}
    \begin{tabular}{c|cccc}
        \toprule
        Method & \#param & SNLI-VE & MUsTARD (A+L) & RefCOCO (val) \\
        \midrule
        LoRA$+P_m$ & 2.1M& 70.77 & 65.54 & 52.15 \\      
        
        BitFit$+P_m$ &  1.9M& 67.54 & 64.34& 53.90\\ 
        
        AdaptFormer$+P_m$ & 3.2M& 69.30 & 65.20 & 52.68 \\   
        \midrule
        MoPE ($k$=4) & 1.6M & \textbf{73.47} &\textbf{ 68.73} & \textbf{58.40} \\
        \bottomrule
    \end{tabular}
    \end{adjustbox}
    \label{tab:peft}
\end{table}

\rb{In this subsection, we compare MoPE with other representative PEFT methods, including AdaptFormer~\cite{chen2022adaptformer}, LoRA~\cite{hu2021lora}, and BitFit~\cite{zaken2022bitfit}. However, a direct comparison is challenging, as these methods are originally designed for unimodal transfer learning rather than multimodal fusion. To facilitate a fair comparison, we adapt these methods to our fusion task by modifying the MoPE architecture: we remove its static and dynamic prompt components and replace them with alternative PEFT modules, while retaining the mapped prompt. We set the bottleneck dimension for AdaptFormer to 64 and the rank for LoRA to 8. The results of this comparison are summarized in Table~\ref{tab:peft}.}

\rb{The results indicate that MoPE achieves superior performance with greater parameter efficiency compared to the other PEFT methods when adapted for multimodal fusion. Unlike these general-purpose PEFT methods, MoPE is specifically designed for PEMF and features an instance-wise conditioning mechanism. This design enables more adaptive fusion, ultimately leading to improved performance on multimodal tasks.}

\subsection{\rb{How MoPE Achieves Greater Scalability}}
\rb{We have demonstrated that the proposed MoPE framework scales more effectively with both trainable parameters and data volume. This enhanced scalability is not achieved by learning new attention patterns, but rather by generating the most effective prompt for each input instance. Specifically, conventional prompt tuning indiscriminately applies the same prompt to all multimodal inputs. This process can be characterized as bottom-up, as it limits the model's expressive power for complex multimodal tasks. In contrast, MoPE generates an adaptive prompt that biases the attention process toward task-relevant tokens. This increased adaptivity enables the primary encoder to perform top-down attention~\cite{anderson2018bottom, lin2024evaluating}, thereby significantly enhancing its expressive power. Another strength of MoPE is its decomposition of the long prompt into multiple specialized experts. This design shifts the challenge from learning a single, universal long prompt~\cite{petrov2023prompting} to optimizing a parameterized router and multiple shorter prompts. Overall, these merits make MoPE a powerful solution for parameter-efficient multimodal fusion.}

\subsection{Limitations and Future Work}
\rb{Despite enhancing empirical expressiveness, the prompts generated by MoPE interact with token embeddings through the same biasing mechanism as conventional prompts. A direct implication is that MoPE cannot induce entirely new attention patterns that were not observed during pre-training~\cite{wang2023universality,petrov2023prompting}. While this characteristic may be advantageous for preserving pre-trained knowledge and preventing catastrophic forgetting, it also limits the scope of downstream applications. This limitation highlights a valuable direction for future research: exploring how cross-modal information can guide the main modality to produce novel attention patterns without training. We believe that this approach could further expand the expressive power of prompt fusion for an even wider range of multimodal tasks.}

\section{Conclusion}
\reva{To the best of our knowledge, we present the first comprehensive study to tackle the long-standing limitations of adaptivity and expressiveness in prompt-based multimodal fusion.  Our central insight involves a divide-and-conquer strategy, leveraging instance-wise adaptive prompts to navigate the problem space. To realize this approach, we have proposed MoPE, which conditions prompt generation on representations derived from all input modalities. Extensive experiments spanning four modalities demonstrate MoPE's superior scalability, surpassing that of vanilla prompt tuning. Furthermore, through regularized routing, we have showcased the high adaptivity and interpretability of MoPE prompting, where the experts spontaneously specialize in different concepts within the multimodal dataset. These strengths make MoPE a highly modular, effective, and efficient method for multimodal fusion, achieving performance comparable to full fine-tuning while utilizing only 0.8\% of the trainable parameters. MoPE's versatility makes it applicable to a wide array of downstream tasks across diverse modalities, which include, but are not limited to RES, VQA, VE, and multimodal sentiment analysis.}

 % \bibliographystyle{IEEEbib}
% \bibliography{main}
\bibliographystyle{IEEEtran}
\bibliography{IEEEabrv,main}


\vspace{-60pt}
\begin{IEEEbiography}[{\includegraphics[width=1in,height=1.25in,clip,keepaspectratio]{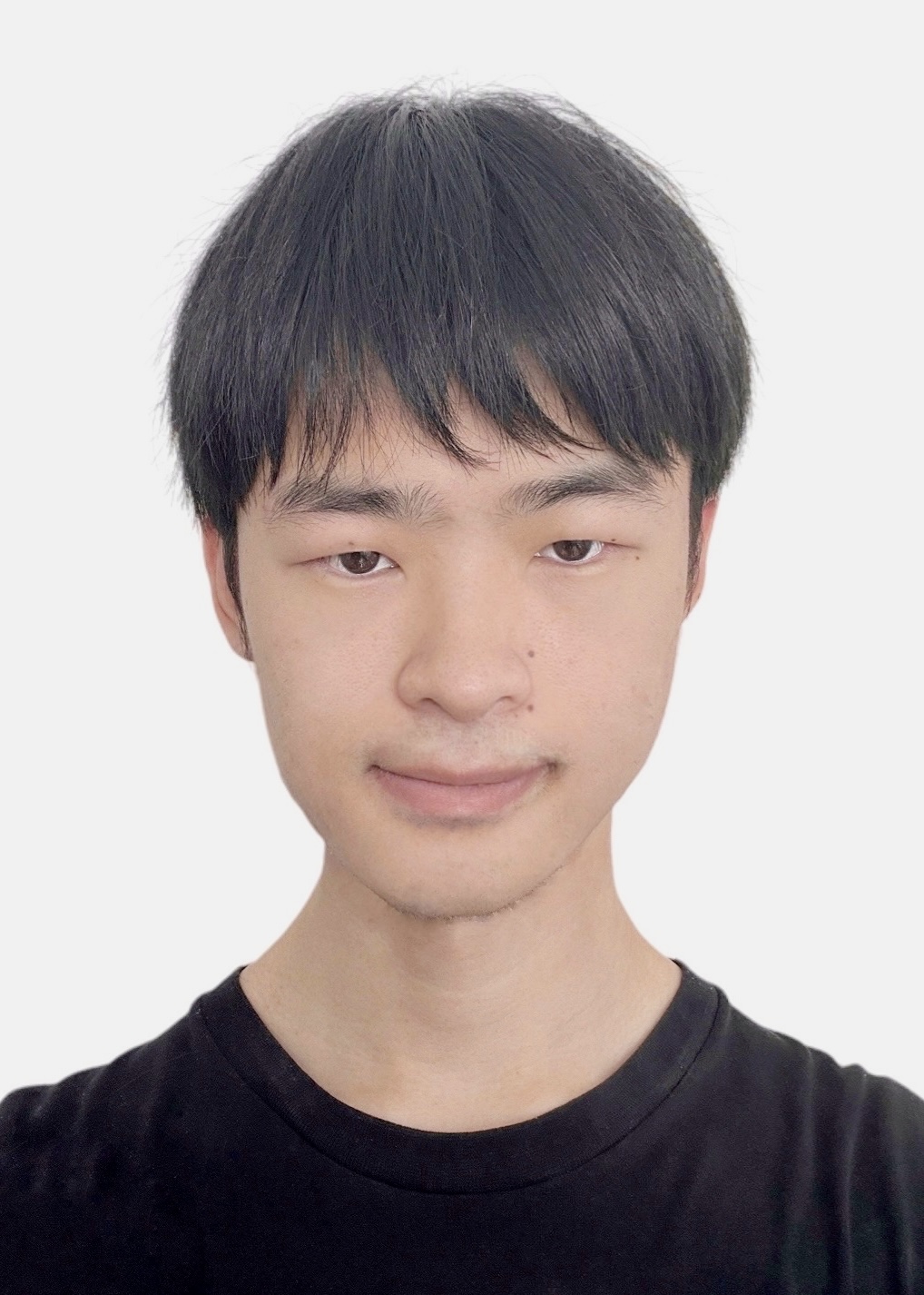}}]{Ruixiang Jiang}
is currently working towards Ph.D. degree at The Hong Kong Polytechnic University (PolyU). He received his B.Sc. degree at PolyU in 2023 with first class honors. His current research interests include computational aesthetics and generative art. He also has broad interests in computer vision, computer graphics, and multimodal learning.
\end{IEEEbiography}

\vspace{-60pt}

\begin{IEEEbiography}[{\includegraphics[width=1in,height=1.25in,clip,keepaspectratio]{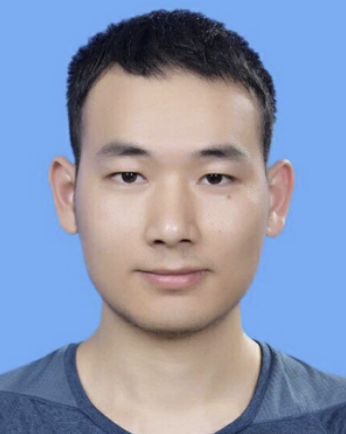}}]{Lingbo Liu}
received the Ph.D. degree in computer
science from the School of Computer Science and
Engineering, Sun Yat-sen University, Guangzhou,
China, in 2020. From 2018 to 2019, he was a
Research Assistant with The University of Sydney,
Camperdown, NSW, Australia. From 2022 to 2024,
he was a Research Assistant Professor with The
Hong Kong Polytechnic University, Hong Kong.
Currently, he is an Associate Researcher with the
Research Institute of Multiple Agents and Embodied Intelligence, Peng Cheng Laboratory, Shenzhen,
China. He has authorized and co-authorized on more than 30 papers in top-tier academic journals and conferences. His research interests include machine
learning, embodied intelligence, and urban computing.
\end{IEEEbiography}

\vspace{-60pt}

\begin{IEEEbiography}[{\includegraphics[width=1in,height=1.25in,clip,keepaspectratio]{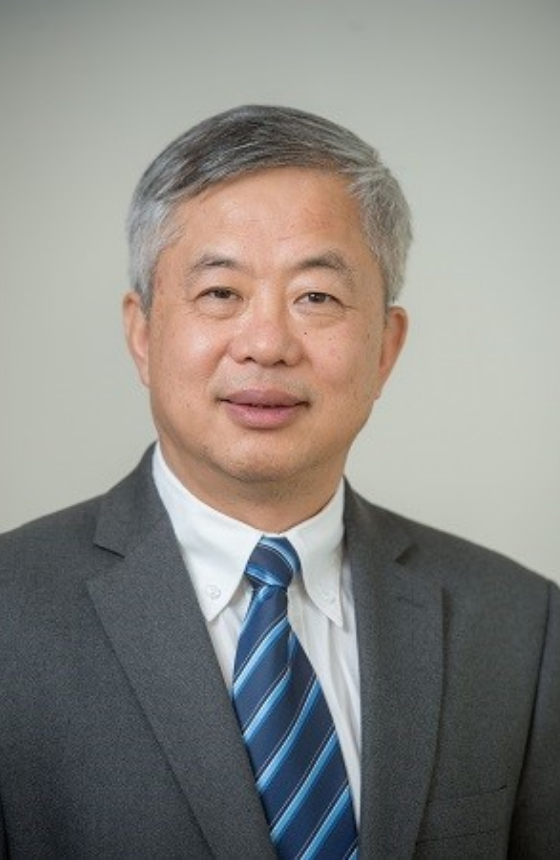}}]{Chang Wen Chen} (Fellow, IEEE)
is currently
Chair Professor of Visual Computing at The Hong Kong Polytechnic University. Before his
current position, he served as
Dean of the School of Science
and Engineering at The Chinese
University of Hong Kong, Shenzhen, from 2017 to 2020, and
concurrently as Deputy Director at Peng Cheng Laboratory
from 2018 to 2021. Previously,
he was an Empire Innovation
Professor at the State University
of New York at Buffalo (SUNY)
from 2008 to 2021 and the Allan
Henry Endowed Chair Professor
at the Florida Institute of Technology from 2003 to 2007. He received
his BS degree from the University of Science and Technology of
China in 1983, an MS degree from the University of Southern California in 1986, and his PhD degree from the University of Illinois
at Urbana-Champaign (UIUC) in 1992.
He has served as Editor-in-Chief for IEEE Trans. Multimedia
(2014-2016) and for IEEE Trans. Circuits and Systems for Video
Technology (2006-2009). He has received many professional achievement awards, including ten (10) Best Paper Awards or Best Student
Paper Awards, the prestigious Alexander von Humboldt Award
in 2010, the SUNY Chancellor’s Award for Excellence in Scholarship and Creative Activities in 2016, the UIUC ECE Distinguished
Alumni Award in 2019, and the ACM SIGMM Outstanding Technical Achievement Award in 2024. He is an IEEE Fellow, a SPIE
Fellow, and a Member of Academia Europaea.
\end{IEEEbiography}

\newpage

\clearpage

\begin{figure*}[t]
\begin{center}
    {\LARGE Appendix for MoPE: Mixture of Prompt Experts for Parameter-Efficient and Scalable Multimodal Fusion} \\[1em]
\end{center}
\end{figure*}

\appendix

\subsection{Architecture Details}\label{sec: more_detail}
This appendix aims to provide additional details on our model architecture and experimental setup.

\textbf{Prompt Vector}. Our prompt implementation closely follows VPT~\cite{jia2022visual}. Specifically, for static prompts and prompt experts, we use uniform initialization $\mathcal{U}\sim (-\eta,\eta)$, where $\eta$ is calculated according to the embedding dimension and patch size of the Transformer~\cite{jia2022visual}. Dropout with $p=0.1$ is applied to all prompts. However, we do not use the reparameterization trick for prompts introduced in the original prompt tuning methods~\cite{li2021prefix}, as the gradients of our dynamic prompt and mapped prompt are already rectified by MLPs (\textit{i.e.,} $\mathbf{W}_x$, $\mathbf{W}_y$, $f_m(\cdot)$). For Transformer architectures that employ a window attention mechanism (\textit{e.g.,} Swin~\cite{liu2021swin}), we duplicate the same prompt and prepend it to all windows for self-attention calculation, following the approach in VPT~\cite{jia2022visual}.

\textbf{Mapper}. We learn a mapper to map representations from the complementary modality $\mathbb{Y}$ to the embedding dimension of the main modality $\mathbb{X}$. Generally speaking, the mapper is implemented as a two-layer MLP with a bottleneck design, which shares similarities with previous work~\cite{li2023efficient}. In our experiments, we set the bottleneck dimension as half of the dimension of the complementary representation, i.e., $d_{bot} = \lceil d_{y}/2 \rceil$. Then, we apply a batch normalization layer and a GeLU activation to obtain a bottleneck feature $\psi_{bot}\in\mathbb{R}_{d_{bot}}$. In practice, we set $d_{bot} = 384$ for the \verb|Swin-base-224| encoder.

\begin{figure}[!t]
    \centering
    \includegraphics[width=\linewidth]{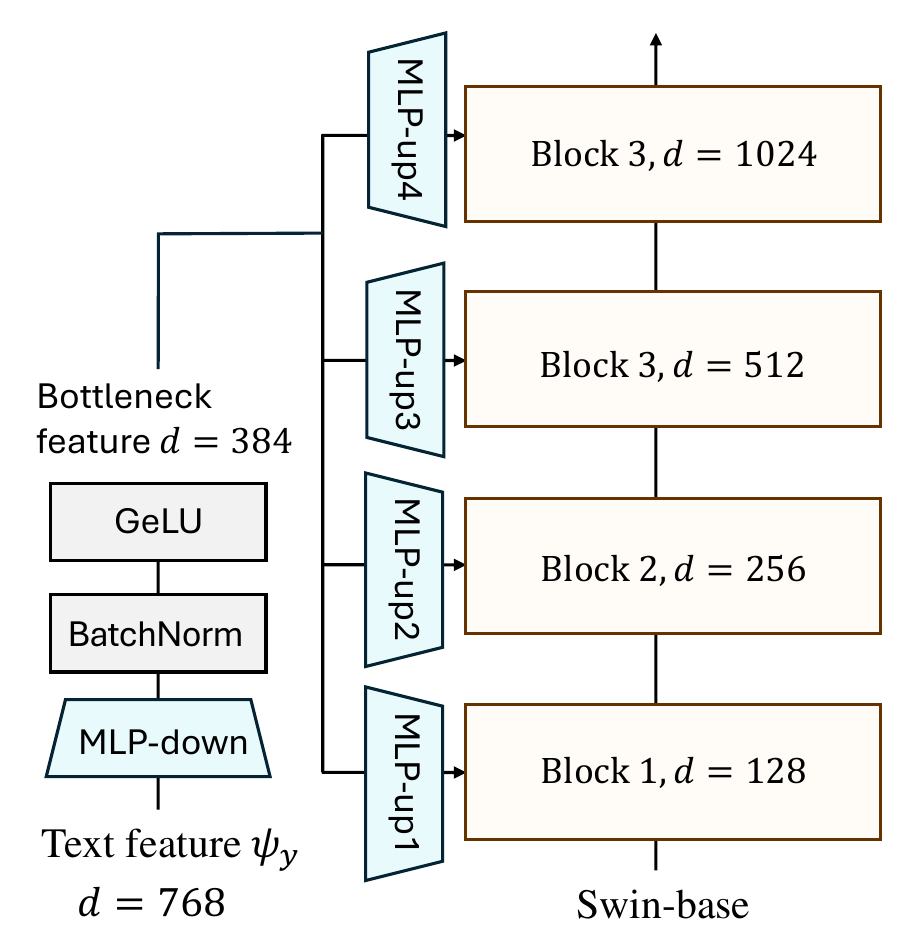}
    \caption{\textbf{Architecture of the Mapper for Swin.} The cross model feature is first projected to a shared low-dimensional bottleneck, then further mapped to different embedding dimensions of each Swin block.}
    \label{fig:mapper}
\end{figure}

After obtaining this bottleneck feature, we apply another linear layer to project it to the dimension of $d_x$. However, some Transformer architectures use inconsistent $d_x$ in different layers. For example, in \verb|Swin-b|~\cite{liu2021swin}, the embedding dimension changes from $[128,256,512,1024]$, making it challenging to fit a single linear transformation. To circumvent this, we instead learn separate up-projections, each used to project the shared bottleneck feature to different embedding dimensions. As a result, there is one single down-sampling layer and multiple up-sampling layers. The design is illustrated in Figure~\ref{fig:mapper}. To ensure fairness, we also use a similar design in plain Transformers such as ViT~\cite{dosovitskiy2020image} and BERT~\cite{devlin2019bert}. 

\textbf{MoPE}. As discussed in the main body of this paper, we learn a per-layer linear projection $\mathbf{W}_x^i$ to obtain the cross-modal embedding. Here, we would like to further clarify that the weight is not shared with the one used in the mapper $f_m$. In our experiments, we use $d_c = 8$ and $d_i = 2$, resulting in $d_r = 10$. For the importance loss, we scale it by a factor of $0.01$ and linearly combine it with the task-specific loss(es) for optimization.

% In \cref{} we show how the relationship of different linear projection in our method.

% \subsection{Dataset Examples}

% In Figure~\ref{fig:dataset_example}, we provide examples of each dataset.

% \begin{figure*}[!t]
%     \centering
%     \includegraphics[width=\linewidth]{figs/fig_dataset_example.pdf}
%     \caption{\textbf{Examples of each dataset used in the main paper.}}
%     \label{fig:dataset_example}
% \end{figure*}

\subsection{Data Processing}\label{sec: data_detail}
 We present details on our data pre-processing in this section. 
 
 For all input images except for the RefCOCO(+), we perform RandAug with $N=2$ and $M=5$, resize them to $(256,256)$, and perform center cropping to obtain images of size $(224, 224)$. 

For the SNLI-VE dataset, we follow Frozen and PMF and use only the image premise, which means that the input to the model is the image premise + text hypothesis. Note that this setting may differ from other works that also use the text premise.
\IEEEpubidadjcol
For the MuSTARD dataset, the overall processing pipeline follows PromptFuse~\cite{liu2022prompt}. For the audio modality, we extract WAV audio from the video using a sampling rate of 16,000 Hz. \rb{For the experiment with PANNs~\cite{kong2020panns} backbone, audio is resampled at 32,000 Hz to maintain consistency with the PANNs pretraining.} We also remove background noise using the 
Librosa package. For a batch of input, we zero-pad the WAV input to the same length. As for the video, our processing is slightly different from PromptFuse~\cite{liu2022prompt}, which uses a face detection model to sample frames where the speaker's face is visible. We instead evenly sample $8$ frames for all videos. The sampled frames are resized to $(224, 224)$ for encoding. In other words, we represent the raw input of a video as a tensor of shape $(8,3,224,224)$. For the text modality, we use the speaker-dependent setting, which means that the speaker's name is visible to the model during both training and inference. To achieve this, we simply concatenate the speaker's name to the input utterance string.

% \subsection{Non-Existence of a Single Global Prompt Achieving the Optimal Bound for All Instances}\label{sec: theory}
\subsection{Theoretical Analysis on MoPE}\label{sec: theory}

This section aims to provide further theoretical analysis on the effectiveness of proposed MoPE design.

Following recent theoretical work on prompt expressiveness~\cite{wang2023universality,petrov2023prompting}, we analyze the scenario for a single prompt (i.e., \(l=1\)) in a single Transformer layer, which lower-bounds the expressiveness of the whole Transformer. For an input \(\mathbf{x}\) and a prompt \(\mathbf{P}\), let the corresponding attention map be denoted as \(\mathcal{A}(\mathbf{x}, \mathbf{P})\) (with the prompt part truncated for simplicity).

For a given task, prompt tuning seeks a prompt \(\mathbf{P}\) such that the induced attention map \(\mathcal{A}(\mathbf{x},\mathbf{P})\) is close to a \emph{task-specific optimal attention pattern} (which minimizes the task loss \(\mathcal{L}(\mathbf{x};\mathbf{P})\)). We define the distance between two attention maps \(\mathbf{A}_1\) and \(\mathbf{A}_2\) as
\[
d(\mathbf{A}_1, \mathbf{A}_2) = \|\mathbf{A}_1 - \mathbf{A}_2\|_{\mathbb{A}},
\] where $||\cdot||_\mathbb{A}$ is a metric in the space $\mathbb{A}$.
% which satisfies the triangle inequality:
% \[
% \|\mathbf{A} - \mathbf{B}\|_{\mathcal{A}} \le \|\mathbf{A} - \mathbf{C}\|_{\mathcal{A}} + \|\mathbf{C} - \mathbf{B}\|_{\mathcal{A}}.
% \]

For each instance \(\mathbf{x}\), denote the set of optimal attention patterns (achieving the best performance for the task) by
% \[
% \mathcal{A}^*(\mathbf{x}) \triangleq \arg\min_{\mathbf{A}\in\mathcal{A}} \|\mathbf{A} - \text{target}(\mathbf{x})\|_{\mathcal{A}}.
% \]
\[
\mathcal{A}^*(\mathbf{x}) \triangleq \arg\min_{\mathbf{A}\in\mathbb{A}}(\mathcal{L}(\mathbf{x})).
\]
Let the minimal error achievable on instance \(\mathbf{x}\) via prompting be defined as
\[
\epsilon^*(\mathbf{x}) \triangleq \inf_{\mathbf{P}\in \mathcal{P}} \Delta(\mathbf{x}, \mathbf{P})\ge 0,
\]
where the error (or discrepancy) for a specific prompt \(\mathbf{P}\) is given by
\[
\Delta(\mathbf{x}, \mathbf{P}) = \inf_{\mathbf{P} \in \mathbb{P}} \|\mathcal{A}(\mathbf{x}, \mathbf{P}) - \mathcal{A}^*(\mathbf{x})\|_{\mathbb{A}}.
\]where $\mathbb{P}$ is the space for all prompts.

By definition, there exists (at least) an instance-specific (optimal) prompt \(\mathbf{P}_\mathbf{x}^*\) achieving
\[
\Delta(\mathbf{x}, \mathbf{P}_\mathbf{x}^*) = \epsilon^*(\mathbf{x}).
\]

\textbf{Assumption.} Assume there exist at least two distinct instances \(\mathbf{x}_1,\mathbf{x}_2\) such that their optimal attention sets are non-overlapping, i.e.,
\[
\delta \triangleq \min_{\substack{\mathbf{A}_1 \in \mathcal{A}^*(\mathbf{x}_1)\\ \mathbf{A}_2 \in \mathcal{A}^*(\mathbf{x}_2)}} \|\mathbf{A}_1 - \mathbf{A}_2\|_{\mathbb{A}} > 0,
\] which is a practical assumption especially for multimodal tasks, where the model need to selectively attend to different sub-sequences during attention (e.g., VQA, RES). We now state the main theorem.
\begin{theorem}[No Global Prompt Achieves Instance-Optimal Error Simultaneously]
Under the above setup and assumption, for any global prompt \(\mathbf{P}_{\text{shared}}\in \mathbb{P}\) (i.e., one that is used for all instances), it holds that
\[
\Delta(\mathbf{x}_1, \mathbf{P}_{\text{shared}}) + \Delta(\mathbf{x}_2, \mathbf{P}_{\text{shared}}) > \epsilon^*(\mathbf{x}_1) + \epsilon^*(\mathbf{x}_2).
\] which implies that the accumulated error with global prompt must be higher than the sum of the instance-specific minimal errors.
\[\sum_{\mathbf{x} \in \mathcal{X}'} \Delta(\mathbf{x}, \mathbf{P}_{\text{shared}}) >\sum_{\mathbf{x} \in \mathcal{X}'}\Delta(\mathbf{x}, \mathbf{P}_\mathbf{x}^*) \]

\end{theorem}

\begin{proof}

Assume, for the sake of contradiction, that there exists a global prompt \(\mathbf{P}_{\text{shared}}\) such that
\[
\Delta(\mathbf{x}_1, \mathbf{P}_{\text{shared}}) + \Delta(\mathbf{x}_2, \mathbf{P}_{\text{shared}}) = \epsilon^*(\mathbf{x}_1) + \epsilon^*(\mathbf{x}_2).
\]
Since for each instance \(\Delta(\mathbf{x}, \mathbf{P}_{\text{shared}}) \ge \epsilon^*(\mathbf{x})\), this equality forces 
\[
\Delta(\mathbf{x}_1, \mathbf{P}_{\text{shared}}) = \epsilon^*(\mathbf{x}_1) \quad \text{and} \quad \Delta(\mathbf{x}_2, \mathbf{P}_{\text{shared}}) = \epsilon^*(\mathbf{x}_2).
\]
In other words, the shared prompt
\(\mathbf{P}_{\text{shared}}\) must simultaneously yield
\[
\mathcal{A}(\mathbf{x}_1, \mathbf{P}_{\text{shared}}) \in \mathcal{A}^*(\mathbf{x}_1)
\]
and 
\[
\mathcal{A}(\mathbf{x}_2, \mathbf{P}_{\text{shared}}) \in \mathcal{A}^*(\mathbf{x}_2).
\]

Which contradicts with the $\delta$-separation of the two optimal attention sets:
\[
\delta \triangleq \min_{\substack{\mathbf{A}_1 \in \mathcal{A}^*(\mathbf{x}_1) \\ \mathbf{A}_2 \in \mathcal{A}^*(\mathbf{x}_2)}} \|\mathbf{A}_1 - \mathbf{A}_2\|_{\mathbb{A}} > 0.
\]
Therefore, no single global prompt can simultaneously achieve the minimal (instance-specific) errors for both instances.

\end{proof}
% Furthermore, by the limited expressiveness of prompt-tuning~\cite{petrov2023prompting}, we have:

% \begin{equation}\Delta(\mathbf{x}_2, \mathbf{P}_2^*) \ge 0; \\ \Delta(\mathbf{x}_1, \mathbf{P}_1^*) \ge 0\end{equation}

% Hence, the following relationship holds:
% \begin{equation}\Delta(\mathbf{x}_1, \mathbf{P}_\text{shared}^*) + \Delta(\mathbf{x}_2, \mathbf{P}_\text{shared}^*) \geq \Delta(\mathbf{x}_1, \mathbf{P}_1^*) + \Delta(\mathbf{x}_2, \mathbf{P}_2^*) \ge 0\end{equation}
\begin{theorem}[Improved Adaptivity of MoPE]
Let 
\[
\mathcal{E} = \{\mathcal{E}_1, \mathcal{E}_2, \ldots, \mathcal{E}_k\}
\]
be a set of \(k\) expert attention mappings, where each 
\(\mathcal{E}_i\colon \mathcal{X} \rightarrow \mathcal{A}\) 
maps an input \(\mathbf{x} \in \mathcal{X}\) to a set of attention patterns \(\mathcal{E}_i(\mathbf{x}) \subseteq \mathcal{A}\). Note that in practice we always have $d>>k$. Define the induced attention mapping of MoPE as:
\[
\mathcal{A}_{\text{MoPE}}(\mathbf{x}, \mathcal{E}, \mathbf{r}) = \left\{\sum_{i=1}^k r_i(\mathbf{x})\, \mathbf{A}_i ~\Bigg|~ \mathbf{A}_i \in \mathcal{E}_i(\mathbf{x}),\, \forall i \right\},
\]
where the routing weights \(\mathbf{r}(\mathbf{x}) \coloneqq [r_1(\mathbf{x}), \ldots, r_k(\mathbf{x})]\) satisfy:
\[
\sum_{i=1}^k r_i(\mathbf{x}) = 1, \quad r_i(\mathbf{x}) > 0.
\]

Let \(\mathcal{X}' \subseteq \mathcal{X}\) be a set of instances. In addition, assume:
\begin{enumerate}
    \item \textbf{(Specialized Experts Condition)} For each instance \(\mathbf{x} \in \mathcal{X}'\), the convex hull of the expert attention mappings,
    \begin{equation*}
        \begin{aligned}
&\operatorname{conv}(\mathcal{E}(\mathbf{x})) \\
&= \Bigg\{ \sum_{i=1}^k \alpha_i\, \mathbf{A}_i ~\Bigg|~ \mathbf{A}_i \in \mathcal{E}_i(\mathbf{x}),\, \forall i,\; \sum_{i=1}^k \alpha_i = 1,\; \alpha_i \ge 0 \Bigg\}
\end{aligned}
\end{equation*}

    contains an optimal attention pattern, i.e.,
    \[
    \mathcal{A}^*(\mathbf{x}) \subseteq \operatorname{conv}(\mathcal{E}(\mathbf{x})), \quad \forall \mathbf{x} \in \mathcal{X}'.
    \]
    
    \item \textbf{(Manifold Assumption)} There exists a manifold \(\mathcal{M}\subset \mathbb{R}^d\) with intrinsic dimension \(m \ll d\) such that for each instance \(\mathbf{x} \in \mathcal{X}'\), the optimal attention pattern \(\mathbf{A}^*(\mathbf{x})\) lies on (or is approximately confined to) \(\mathcal{M}\), and 
    \[
    \mathcal{M} \subseteq \operatorname{conv}(\mathcal{E}(\mathbf{x})).
    \]
\end{enumerate}

Then, there exists an instance-dependent routing function \(\mathbf{r}^*\colon \mathcal{X} \to \Delta^{k-1}\) such that the accumulated attention discrepancy for MoPE under \(\mathbf{r}^*\) across instances in \(\mathcal{X}'\) is equal to the sum of the optimal instance-wise attention discrepancies, i.e.,
\[
\sum_{\mathbf{x} \in \mathcal{X}'} \Delta(\mathbf{x}, \mathcal{E}, \mathbf{r}^*) = \sum_{\mathbf{x} \in \mathcal{X}'} \inf_{\mathbf{A} \in \mathcal{A}^*(\mathbf{x})} \|\mathbf{A} - \mathbf{A}^*\|_\mathbb{A}.
\]
\label{theory:2}
\end{theorem}

\begin{proof}
We consider two cases:

\textbf{Case 1:} \( |\mathcal{X}'| \le k \). \\
In this case, the proof is trivial since one may “store” the optimal prompt for each instance in one or a few of the experts.

\textbf{Case 2:} \( |\mathcal{X}'| > k \). \\
Let \(\mathcal{X}' \subseteq \mathcal{X}\) be a set of instances with cardinality exceeding the number of experts. By the \textbf{Specialized Experts Condition} and the \textbf{Manifold Assumption}, for each instance \(\mathbf{x} \in \mathcal{X}'\) the optimal attention mapping \(\mathbf{A}^*(\mathbf{x})\) lies in the convex hull of the expert mappings:
\[
\mathbf{A}^*(\mathbf{x}) \in \operatorname{conv}(\mathcal{E}(\mathbf{x})).
\]

Since the routing weights \(\mathbf{r}(\mathbf{x})\) form convex combinations of the expert outputs, the induced attention mapping of MoPE satisfies:
\[
\mathcal{A}_{\text{MoPE}}(\mathbf{x}, \mathcal{E}, \mathbf{r}) = \operatorname{conv}(\mathcal{E}(\mathbf{x})), \quad \forall \mathbf{x} \in \mathcal{X}'.
\]

Thus, for each instance \(\mathbf{x} \in \mathcal{X}'\) there exists an attention pattern 
\[
\mathbf{A}^* \in \mathcal{A}^*(\mathbf{x}) \cap \mathcal{A}_{\text{MoPE}}(\mathbf{x}, \mathcal{E}, \mathbf{r}^*)
\]
for some instance-dependent routing function \(\mathbf{r}^*\). By construction, the attention discrepancy for each instance under \(\mathbf{r}^*\) is
\begin{equation*}
    \begin{aligned}
\Delta(\mathbf{x}, \mathcal{E}, \mathbf{r}^*) &= \inf_{\mathbf{A} \in \mathcal{A}_{\text{MoPE}}(\mathbf{x}, \mathcal{E}, \mathbf{r}^*)} \|\mathbf{A} - \mathbf{A}^*\|_\mathbb{A} \\
&= \inf_{\mathbf{A} \in \mathcal{A}^*(\mathbf{x})} \|\mathbf{A} - \mathbf{A}^*\|_\mathbb{A}, \quad \forall \mathbf{x} \in \mathcal{X}'. 
    \end{aligned}
\end{equation*}

Therefore, the accumulated attention discrepancy under \(\mathbf{r}^*\) is given by:
\[
\sum_{\mathbf{x} \in \mathcal{X}'} \Delta(\mathbf{x}, \mathcal{E}, \mathbf{r}^*) 
=\sum_{\mathbf{x} \in \mathcal{X}'} \inf_{\mathbf{A} \in \mathcal{A}^*(\mathbf{x})} \|\mathbf{A} - \mathbf{A}^*\|_\mathbb{A},
\]
as required.
\end{proof}

Putting it together, the Theorem.~1 state that the global-shared prompt (\textit{i.e.,} vanilla prompt) could not achieve the best result, when the input instances require different optimal attention patterns. Theorem.~2 state the possibility for MoPE to achieve the optimal result on all of the input instances, conditioned on the appropriate specialization of experts.

% \section*{Additional Visualizations}
% \label{sec:broader}
% In this appendix, we provide extended visualizations of MoPE, including qualitative comparisons of RES and routing visualizations.

\subsection{Visualization of Referring Expression Segmentation Results}
\begin{figure*}
    \centering
    \includegraphics[width=\linewidth]{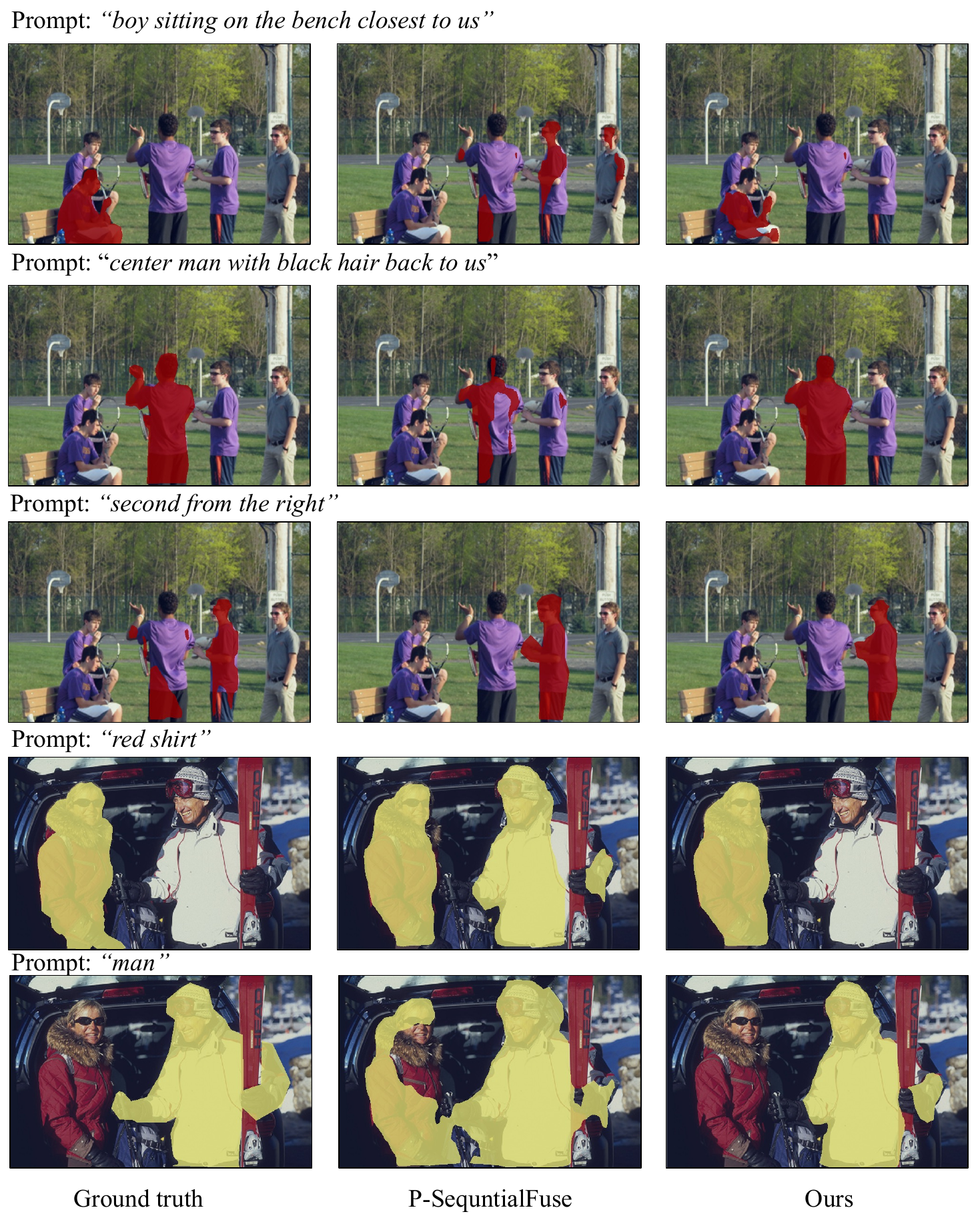}
    \caption{\textbf{Visualization on Referring Expression Segmentation - 1}}
    \label{fig:seg_1}
\end{figure*}

\begin{figure*}
    \centering
    \includegraphics[width=\linewidth]{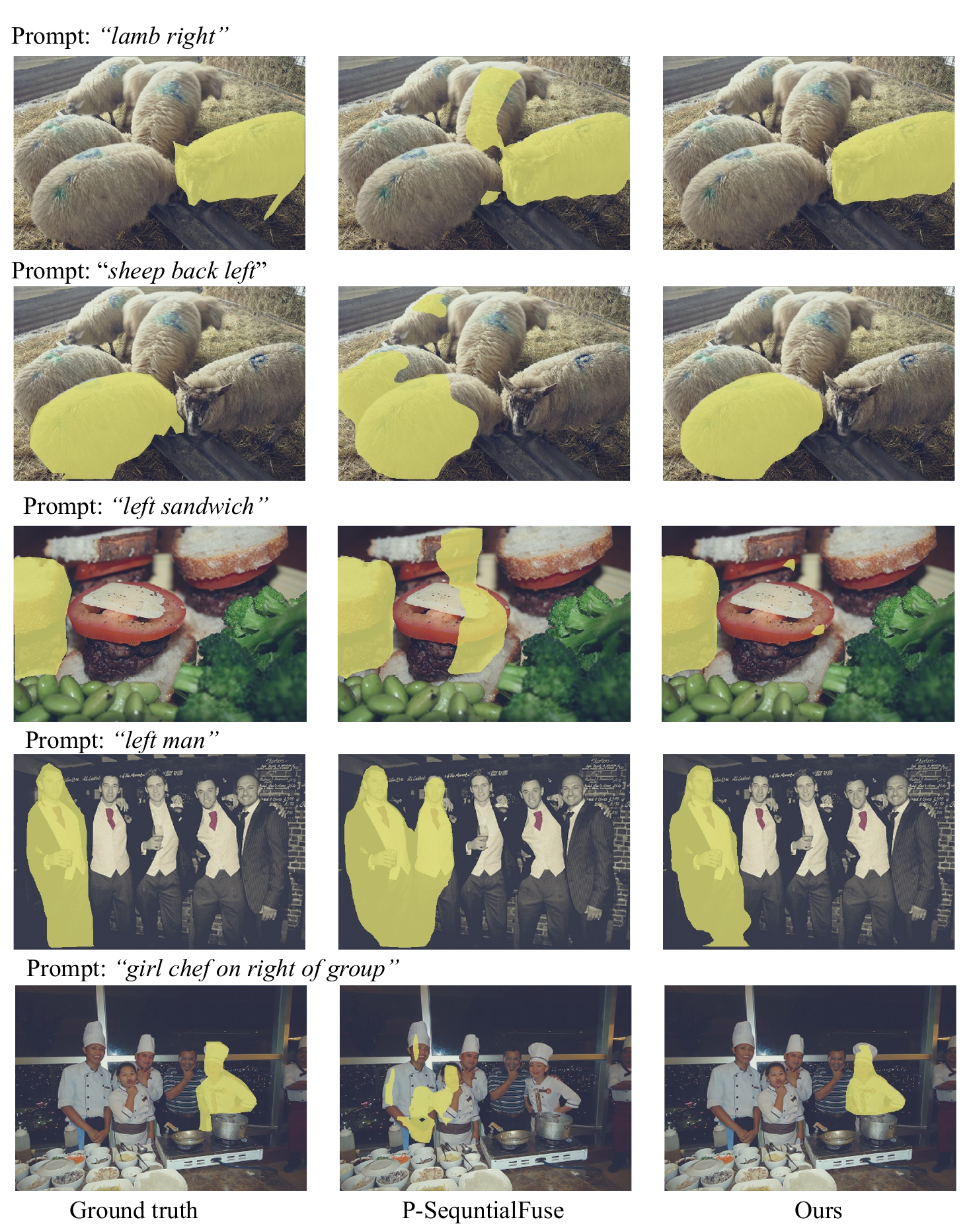}
    \caption{\textbf{Visualization on Referring Expression Segmentation - 2}}
    \label{fig:seg_2}
\end{figure*}

In Figure~\ref{fig:seg_1}, Figure~\ref{fig:seg_2} we provide a qualitative comparison on our method and the \textit{P-SeqFuse} baseline. As illustrated in the figures, our methods could correctly understand the referring expression and localize the object mask accordingly. By contrast, the compared method may fail to follow the text guidance, leading to less accurate segmentation results.

\subsection{\rb{Additional Examples of Expert Routing on VQAv2}}
\label{sec:vqa}

\begin{figure*}
    \centering
    \includegraphics[width=0.8\linewidth]{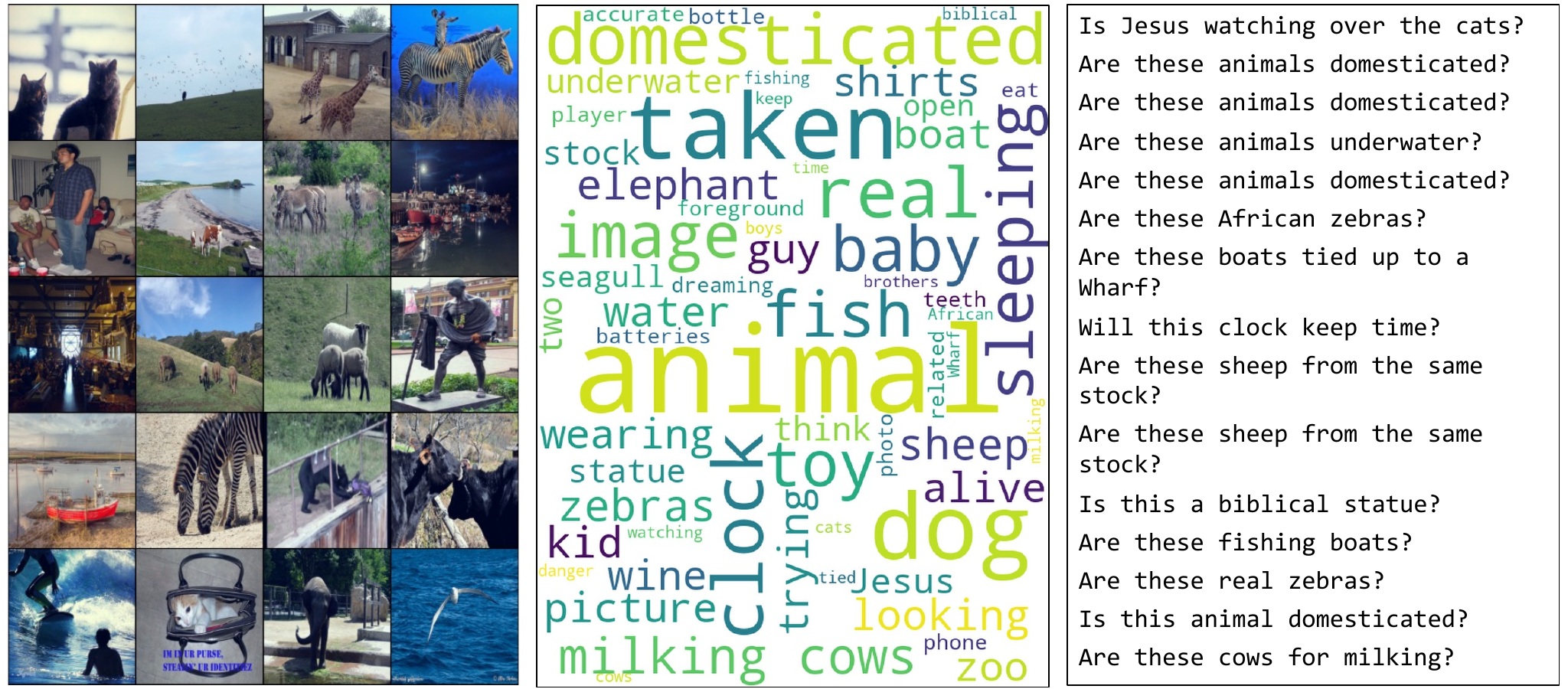}
    \caption{\rb{\textbf{Additional example of expert routing - 1.} We show routing result of expert-4 on the VQAv2 dataset, which specialize in animals. Note: Sorted by routing score, images and texts may not be in the same order.}}
    \label{fig:more_route_1}
\end{figure*}

\begin{figure*}
    \centering
    \includegraphics[width=0.8\linewidth]{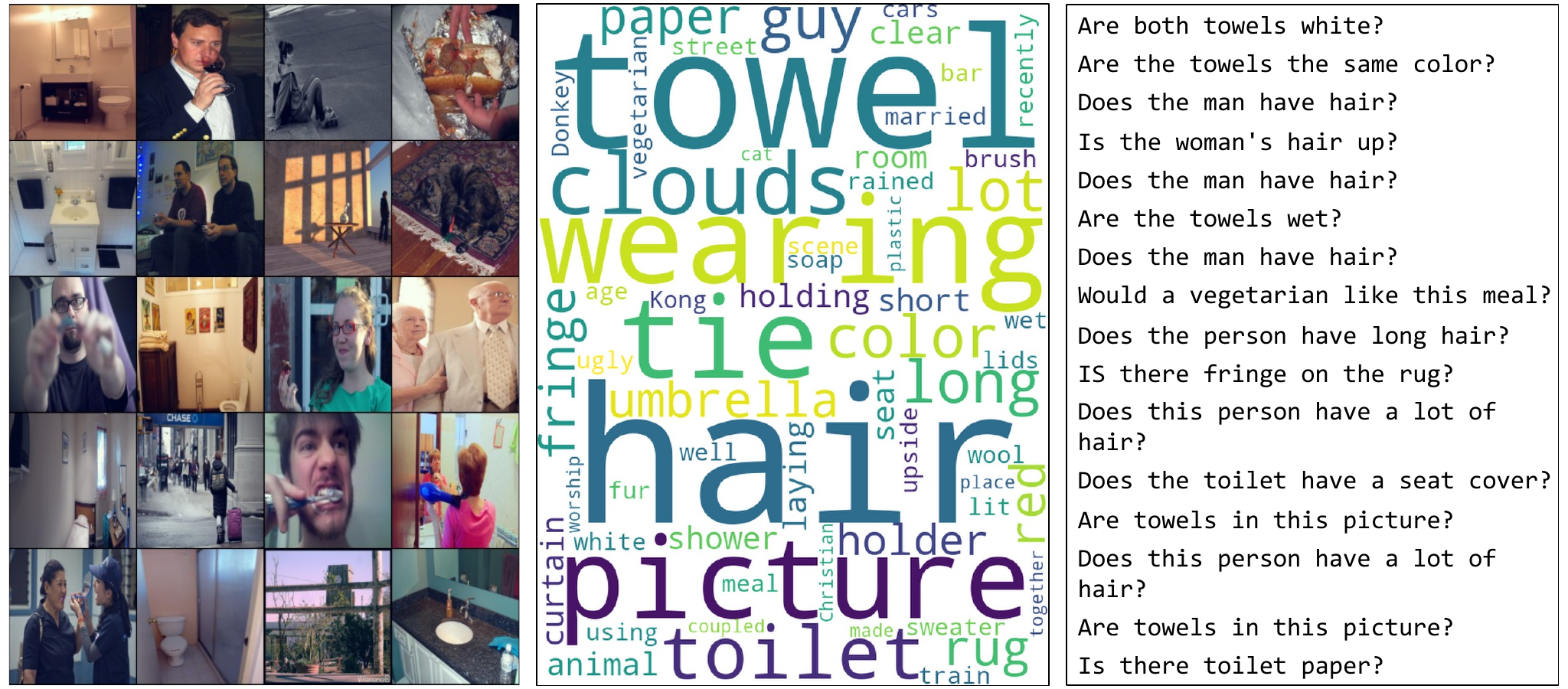}
    \caption{\rb{\textbf{Additional example of expert routing - 3.} We show routing result of expert-8 on the VQAv2 dataset, which specialize in toilet-related concepts and hairstyles. Note: Sorted by routing score, images and texts may not be in the same order.}}
    \label{fig:more_route_3}
\end{figure*}

\begin{figure*}
    \centering
    \includegraphics[width=0.8\linewidth]{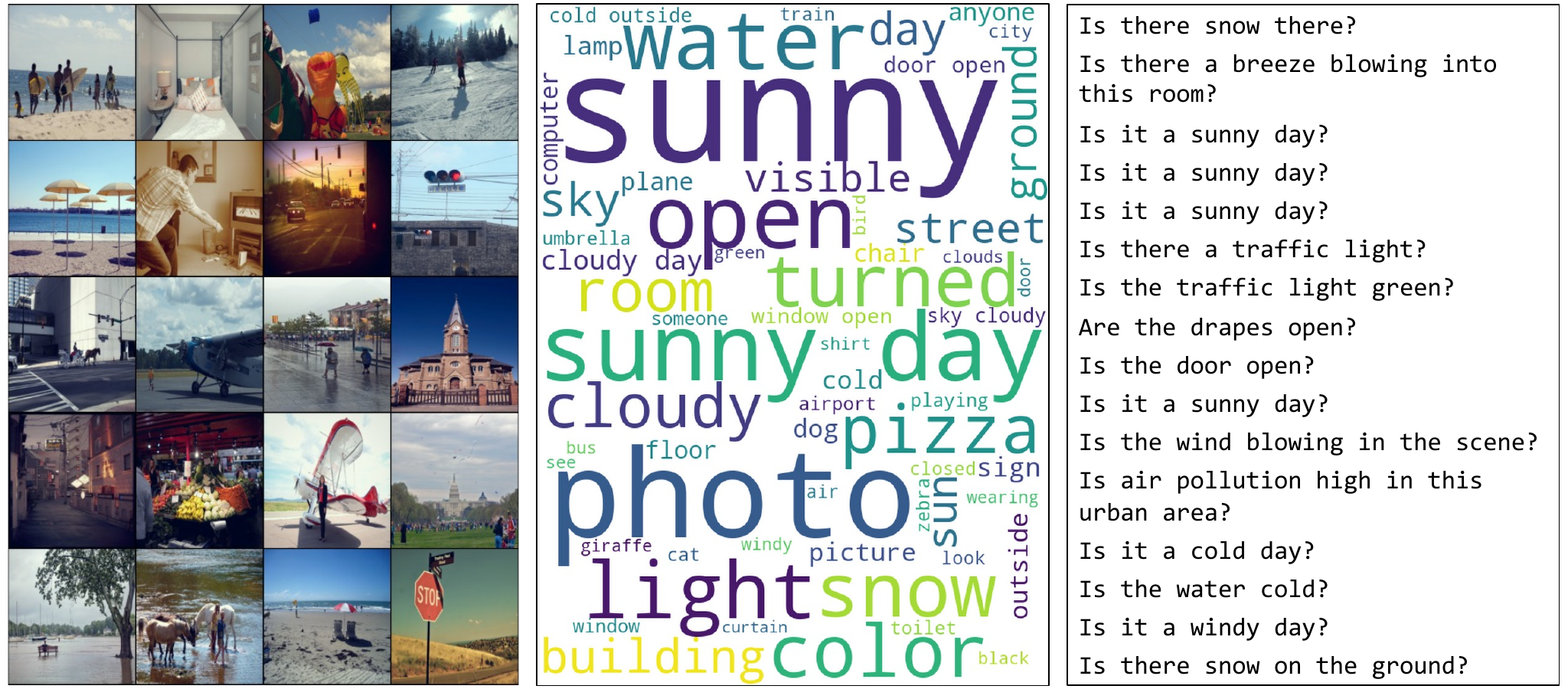}
    \caption{\rb{\textbf{Additional example of expert routing - 4.} We show routing result of expert-10 on the VQAv2 dataset, which specialize in concept related weather conditions. Note: Sorted by routing score, images and texts may not be in the same order.}}
    \label{fig:more_route_4}
\end{figure*}

\rb{To further illustrate the learned concepts via expert specialization, we train our MoPE-based method on an even larger dataset, VQAv2. This dataset contains 265,016 images and paired questions, covering various visual and textual concepts. We visualize the routing results the same way as in Figure~3 of the main text. The results are presented in Figures~\ref{fig:more_route_1}, \ref{fig:more_route_3}, and~\ref{fig:more_route_4}. These examples clearly show specialization, where different experts capture different concepts.}


\end{document}